\documentclass{article}

\PassOptionsToPackage{numbers}{natbib}

\usepackage[final]{neurips_data_2022}

\usepackage[utf8]{inputenc} 
\usepackage[T1]{fontenc}    
\usepackage{hyperref}       
\usepackage{url}            
\usepackage{booktabs}       
\usepackage{amsfonts}       
\usepackage{nicefrac}       
\usepackage{microtype}      
\usepackage{xcolor}         

\usepackage{graphicx}
\graphicspath{{Figures/}}\usepackage{float}
\usepackage{subfig}
\usepackage{amsmath}
\usepackage{multirow}

\RequirePackage{xspace}

\makeatletter
\DeclareRobustCommand\onedot{\futurelet\@let@token\@onedot}
\def\@onedot{\ifx\@let@token.\else.\null\fi\xspace}

\def\eg{\emph{e.g}\onedot} 
\def\ie{\emph{i.e}\onedot}

\makeatother

\newcommand{\DetectorNetwork}{f}

\title{MBW: Multi-view Bootstrapping in the Wild}

\author{Mosam Dabhi$^{1}$ \hspace{-0.35cm}
\And Chaoyang Wang$^{1}$ \hspace{-0.38cm}
\And Tim Clifford$^{2}$ \hspace{-0.38cm} 
\And L{\'a}szl{\'o} A. Jeni${^{1}}$\vspace{-0.5em}{\hspace{-0.05cm}\thanks{indicates the authors advised equally}} \hspace{-0.25cm} 
\AND Ian Fasel${^{2*}}$
\hspace{-9.8cm} 
\And Simon Lucey${^{3*}}$  \\
\vspace{-0.49cm}
\AND $^{1}$ Carnegie Mellon University \hspace{-0.6cm}
\And $^{2}$ Apple \hspace{-0.6cm}
\And $^{3}$ The University of Adelaide 
}

\begin{document}

\maketitle

\begin{abstract}
Labeling articulated objects in unconstrained settings have a wide variety of applications including entertainment, neuroscience, psychology, ethology, and many fields of medicine. Large offline labeled datasets do not exist for all but the most common articulated object categories (e.g., humans). Hand labeling these landmarks within a video sequence is a laborious task. Learned landmark detectors can help, but can be error-prone when trained from only a few examples. Multi-camera systems that train fine-grained detectors have shown significant promise in detecting such errors, allowing for self-supervised solutions that only need a small percentage of the video sequence to be hand-labeled. The approach, however, is based on calibrated cameras and rigid geometry, making it expensive, difficult to manage, and impractical in real-world scenarios. In this paper, we address these bottlenecks by combining a non-rigid 3D neural prior with deep flow to obtain high-fidelity landmark estimates from videos with only two or three uncalibrated, handheld cameras. With just a few annotations (representing 1-2\% of the frames), we are able to produce 2D results comparable to state-of-the-art fully supervised methods, along with 3D reconstructions that are impossible with other existing approaches. Our Multi-view Bootstrapping in the Wild (MBW) approach demonstrates impressive results on standard human datasets, as well as tigers, cheetahs, fish, colobus monkeys, chimpanzees, and flamingos from videos captured casually in a zoo. We release the \href{https://github.com/mosamdabhi/MBW}{\textcolor{magenta}{codebase}} for MBW as well as this challenging \href{https://github.com/mosamdabhi/MBW-Data}{\textcolor{magenta}{zoo dataset}} consisting image frames of tail-end distribution categories with their corresponding 2D, 3D labels generated from minimal human intervention.
\end{abstract}

\section{Introduction} \label{sec: introduction}

Hand labeling landmarks of articulated objects within video is an arduous and expensive task. Landmark detectors~\cite{hrnet,stacked_hourglass,cpm} can be employed to automate the process. However, they require the ingestion of large amounts of labeled training data to be reliable -- an infeasible requirement for all but the most common of articulated objects (e.g. people, hands). Semi-supervision can help~\cite{simon_tomas_hand}, where a small portion of frames within the video are hand labeled. Candidate labels can be generated from the noisy landmark detectors -- trained from the seed hand labeled examples -- inliers are then determined through calibrated rigid multi-view geometry. These inliers are treated as labels and used to train the next round of landmark detectors. This semi-supervised process is iterated to increase the number of inlier estimates, with additional human annotation being added judiciously to ensure the full sequence is labeled. Such strategies have been instrumental for obtaining reliable ground-truth -- most notably the Multi-view Bootstrapping (MB) approach of~\citet{simon_tomas_hand}. Human annotators are only required to hand label a subset of the dataset, with the rest just requiring visual inspection to validate the accuracy of the inferred labels.  

Although significantly cutting down on human labor, Multi-view Bootstrapping~\cite{simon_tomas_hand} is still expensive and cumbersome, requiring a static multi-camera rig which usually consists of tens~\cite{openmonkeystudio} or even sometimes hundreds of calibrated cameras~\cite{panoptic}. The number of cameras can be reduced, but with a trade-off in decreasing robustness of outlier rejection and increasing human interventions (see Fig.~\ref{Fig: Outlier_rejection}). This makes it less feasible for capturing objects outside laboratories. In this paper, we advocate for a significant advancement by enabling its application to data captured by a few (2 to 4) handheld cameras with only a handful of annotated frames (about 10-15 frames per several minutes of video). We refer to our approach herein as~\emph{Multi-view Bootstrapping in the Wild} (MBW). The cameras need not be calibrated, and fields of view need only overlap the articulated object, not the backgrounds. 

Our innovations come from (i) utilizing Multi-View Non-Rigid Structure from Motion (MV-NRSfM)~\cite{dabhi3dv} to more reliably estimate camera poses and 3D landmark positions from noisy 2D inputs with few cameras. Compared to performing SfM / triangulation independently for each frame as in prior works~\cite{panoptic,openmonkeystudio}, MV-NRSfM leverages the redundancy in shape variations among different frames, thus it is less sensitive to the variations of input views, more capable of detecting outliers and denoising inlier 2D landmark estimates.
(ii) We leverage recent advances in deep optical flow~\cite{raft} as an alternative strategy for creating landmark label candidates -- something especially useful in the early iterations of the semi-supervision process.

\begin{figure}[!t]
  \centering
  \includegraphics[width=1\linewidth]{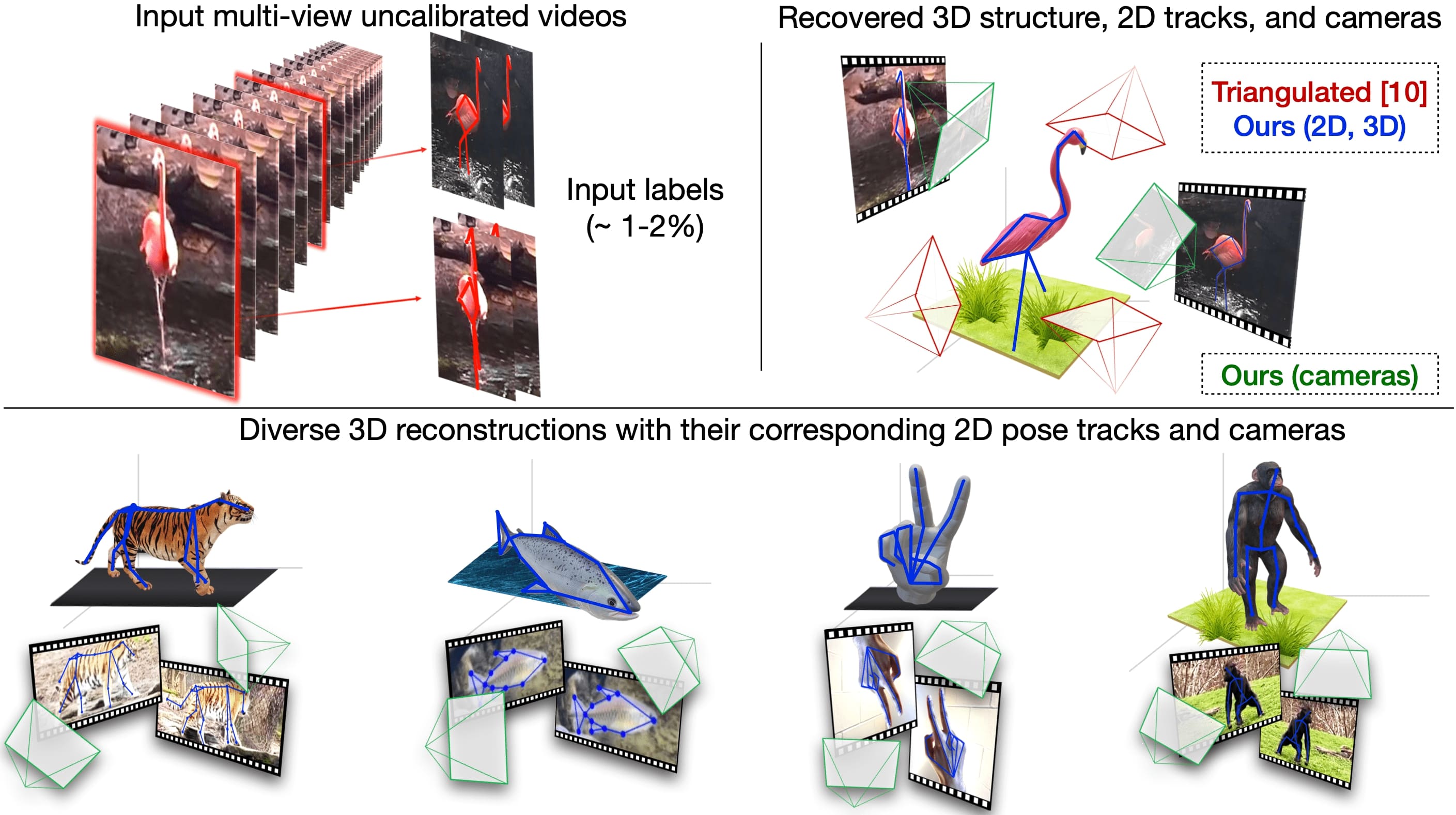}
  \caption{Overview of our MBW approach. \textbf{Top}: Provided an unconstrained Multi-view uncalibrated video with very few 2D labels ($ \approx 1-2$ \% or $\approx15$ labels), our method recovers the 3D structure in a canonical frame, along with  camera poses and corresponding 2D landmarks for the complete video sequence. \textbf{Bottom}: Diverse reconstructions and data labeling for videos captured in the wild. This dataset is released as part of the paper.
  } 
  \label{Fig:intro} 
  \vspace{-0.4cm}
\end{figure}

As a result our approach can be effectively applied to less studied articulated object categories. We show results on tigers, fish, colobus monkeys, gorillas, chimpanzees, and flamingos from a zoo dataset (captured by the authors, who hereby release it under a CC-BY-NC license). We also quantitatively evaluate the proposed pipeline on common motion capture datasets (\eg Human3.6 Million~\cite{human36m}). The accuracy of the learned landmark detector is competitive to state-of-the-art fully supervised method.  A graphical depiction of our approach can be found in Figure~\ref{Fig:intro}.

\section{Related Works} \label{sec: related_works}
Panoptic Studio~\cite{panoptic} paved the way for collecting data for deformable objects such as the human body. Subsequent efforts on humans~\cite{human36m, humanskiing, humanmartial}, hands~\cite{freihand, contrahand, InterHand}, monkeys~\cite{openmonkeystudio}, canines~\cite{dogs_dataset}, cheetahs~\cite{acinoset}, rats~\cite{rats-large-scale}, and insects~\cite{deepfly3d} have followed. Multi-view Boostrapping~\cite{simon_tomas_hand} has demonstrated how these calibrated multi-camera datasets can be labeled efficiently through a semi-supervised learning paradigm and a small number of hand annotations. A fundamental drawback to multi-view bootstrapping however is that it requires a large number of views and accurate camera calibration.

Recent works have explored alternate paradigms for semi-supervised landmark labeling that do not require such exotic calibrated multi-camera setups. ~\citet{deeplabcut}, Pereira et al.~\cite{sleap}, and~\citet{ap10k} tackle this problem from a single view, but largely ignore the use of multi-view geometry. Gunel et al.~\cite{deepfly3d} have explored an approach that utilizes a small number of camera views, and only requires an approximate estimate of the camera extrinsics. They use pictorial structures~\cite{pictorial_structure} to automatically detect and correct labeling errors, and use active learning to iteratively improve landmark detection performance. Although this approach is useful in lab settings where there are static cameras and the object is anchored to a fixed location (e.g. tethered flies are positioned over a spherical treadmill ~\cite{deepfly3d}), it is non-trivial to generalize such performance to more complex environments and across significant individual variations due to e.g.~patterned skins in animals or demographics and clothing in humans. In contrast, our approach accepts image frames from moving cameras and requires only a handful of hand annotated labels. Further, it does not require any camera information, and can easily be applied to a broad set of articulated objects such as humans, hands, and animals. Thus, the strength of our method is its generalizability. Since the provided implementation of DeepFly3D was specific for Drosophila, it was not readily applicable to our in-the-wild datasets. An overview highlighting major differences between our proposed approach and related works trying to achieve a similar application is shown in Tab.~\ref{related-works-table}.

\begin{table}
  \caption{Related efforts trying to achieve a similar application as the proposed approach.}
  \label{related-works-table}
  \centering
  \begin{tabular}{cccccc}
    \toprule
    \cmidrule(r){1-2}
    Method     & Flow     & Calibration & 3D labels & Wild setup & \% annotated ($\approx$) \\
    \midrule
    ~\citet{deepfly3d} & No  & Required & \textbf{Yes} & No &  $ 30\%$\\
    ~\citet{deeplabcut}& No  & N/A & No & No & $  5\%$\\
    ~\citet{sbr}      & \textbf{Yes}  & Required & No & No & N/A (Unknown)\\    
    ~\citet{msbr}      & \textbf{Yes}  & Required & No & No & $ 4\%$\\
    ~\citet{sleap}      & No  & N/A & No & No & $ 5\%$ \\
    ~\citet{simon_tomas_hand}      & No  & Required & \textbf{Yes} & No & $ 30\%$ \\
    \hspace{0.1cm}\textbf{MBW (Ours)}  & \textbf{Yes}  & \textbf{No} & \textbf{Yes} & \textbf{Yes} & $  \mathbf{2}\% $\\
    \bottomrule
  \end{tabular}\vspace{-0.2cm}
\end{table}


\section{Approach}

\begin{figure} [!t]
    \centering
    \includegraphics[width=1\linewidth]{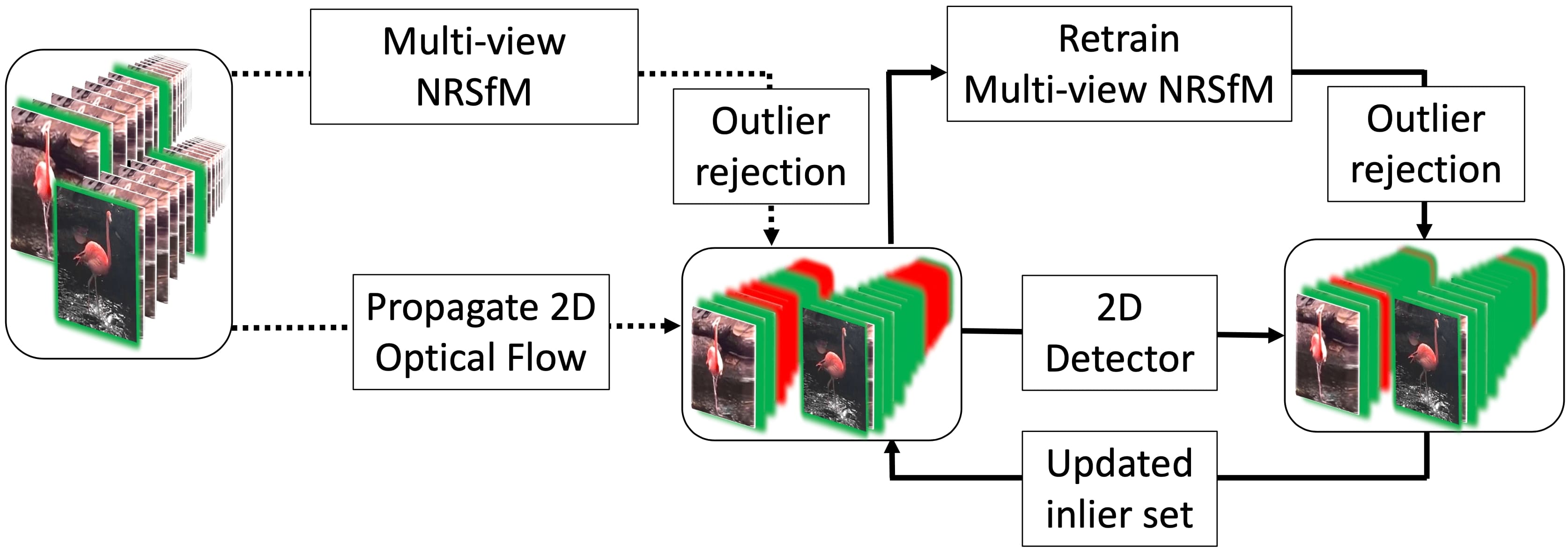}
  \caption{(Dotted lines) The MV-NRSfM neural shape prior is initially trained with labels for 1-2\% of the frames (shown as \textcolor{green}{green} images). A pre-trained optical flow network then propagates the initial labels through the video to generate additional 2D candidates. Candidates that result in high reprojection error from the 3D lifting network are rejected as outliers (\textcolor{red}{red}). (Solid line) From here on, the label set is updated with inliers from the previous iteration, and is then used both to retrain the MV-NRSfM and to train a 2D detector. Dotted line is executed only once while solid lines are repeated for $K$ iterations.} \vspace{-0.5cm}
  \label{Fig: overview_v2}
  \vspace{-0.5cm}
\end{figure}

\subsection{Problem Setup}
Our goal is to learn 2D landmarks of articulated objects from multi-view synchronized videos captured in the wild. Unlike other works~\cite{panoptic,deepfly3d,sbr,msbr} developed for laboratory settings, we focus on the \emph{in the wild} setting, \ie data is captured using a small number (2 or 3) of cameras with \emph{unknown} extrinsics, and only a small portion (1 to 2$\%$) of the data is manually labeled. 

\begin{figure} [!t]
    \centering
    \includegraphics[width=1\linewidth]{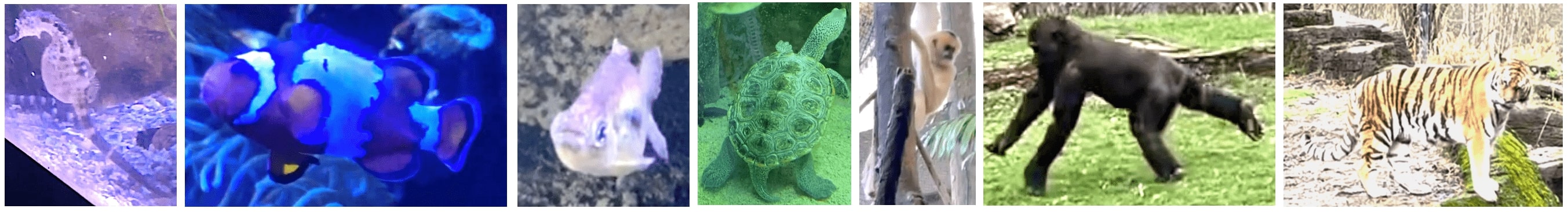}
    \caption{Sample sequences composited from our Zoo data collection -- situations where traditional SLAM pipelines fail to recover reasonable camera matrices due to lack of reliable matching features. 
    }
    \label{fig:dataset_variation}
    \vspace{-0.1cm}
\end{figure}

More specifically, our training set $\mathcal{S}$ consists of $V$ synchronized videos, each with $N$ frames. Each training image is denoted as $\mathbf{I}_{(n,v)}$ where $n \in [1, \ldots, N]$ and $v \in [1, \ldots, V]$ denote frame and view indices. Initially only a subset of frames $(n,v)\in\mathcal{S}_0$ are given with 2D landmark annotations $\mathbf{W}_{(n,v)}\in\mathbb{R}^{P\times2}$ of $P$ points. Each row of $\mathbf{W}_{(n,v)}$ corresponds to the 2D location of a landmark (\eg the left knee of flamingo, see Fig.~\ref{fig:dataset_variation}). To simplify explanations, we assume that only a single object of interest is visible in each frame. For multiple non-overlapping objects, our algorithm is able to estimate bounding boxes to reduce the problem into a single object case (see Appendix~\ref{appendix: bbox}). Finally, the goal is to (i) infer the missing 2D landmark annotations in the training set as a self-labeling task; (ii) train a 2D landmark detector for unseen objects of the same category. 

\subsection{Learnable geometric supervised self-training}
We employ a self-training approach which iteratively assigns pseudo labels and retrains a 2D landmark detector. At each iteration, the 2D pseudo labels generated by a landmark detector are verified using geometric constraints. Samples which fail the verification are dropped, and the remaining pseudo labels are denoised before feeding them back as labels to retrain the landmark detector. Such geometric supervised self-training strategy has been widely used in learning landmark detections~\cite{deepfly3d,panoptic,openmonkeystudio,simon_tomas_hand}, what differentiates our work is that we model the geometric constraints as a \emph{learnable} function, which is learned together with the landmark detector. We abstract this function as:
\begin{equation}
    g: \tilde{\mathbf{W}}_1,\tilde{\mathbf{W}}_2,\ldots,\tilde{\mathbf{W}}_V \longrightarrow y_1,y_2,\ldots,y_V ,  
\end{equation}
where $\tilde{\mathbf{W}}_v\in\mathbb{R}^{P\times2}$ represents detected 2D landmarks at $v$-th view, and $y_v$ is the measured uncertainty for outlier rejection. We derive $g$ from performing multi-view non-rigid structure from motion (MV-NRSfM) as described in Sec.~\ref{sec: neural_prior_outlier}. The remaining details of the self-training pipeline is given as follows.\\

\vspace{-0.6cm}
\paragraph{Initialization} In the initial step, we require human labelers to annotate the 2D landmark positions of the same target object for a small portion of captured video frames. We then train our geometric constraint function $g$ using these initial labels. Since the initial labels only cover a limited range of shape variations, the learned $g$ is aggressive in detecting outliers at the beginning stage of the training. It will be improved as it sees more shape variation in each iteration.\\

\vspace{-0.6cm}
\paragraph{Label propagation through tracking} We find that directly training a 2D landmark detector such as HRNet~\cite{hrnet} using very few labeled samples yields unstable results. To increase the number of training samples, we propagate the annotated 2D landmark labels to the rest of unlabeled frames through tracking. We use an off-the-shelf optical flow network~\cite{raft} to track the landmarks frame to frame. Other tracking methods~\cite{sand2008particle,harley2022particle} can also be utilized. We employ standard forward\&backward flow consistency check to detect tracking failures. Since the optical flow network tends to make consistent wrong estimations when swapping the input frames, such consistency check alone is not enough to exclude all tracking failures. Therefore, we further employ the learned geometric constraint function $g$ to aggressively remove any likely outliers if the predicted uncertainty $y$ is above a certain threshold. We then add the remaining tracked points (inliers) to the labeled set. This new set is then used both to re-train $g$, and to train the first iteration of the 2D landmark detector used in the subsequent stages. 

\vspace{-0.1cm}
\paragraph{Self-training iterations} At each iteration $t$, we define a ``labeled'' set $\mathcal{S}_{t-1}$ which includes all frames that are either manually annotated, or are labeled by the landmark detector $\DetectorNetwork_{t-1}$ in the previous stage and passes outlier rejection using $g_{t-1}$. We then re-train the landmark detector and the geometric constraint function on the labeled set $\mathcal{S}_{t-1}$, which leads to a new detector $\DetectorNetwork_t$ as well as $g_t$. Once trained, inference is run with this detector network $\DetectorNetwork_t$ over all the captured frames. This produces new pseudo labels $\tilde{\mathbf{W}}^t_{n,v}$ for all the $N$ frames and $V$ views. We then apply the geometric constraint function $g_t$ to evaluate the uncertainty score $y^t_{n,v}$ for each pseudo label. Finally we define a new labeled set $\mathcal{S}_t$ which includes all samples $(n,v)$ that satisfy $y^t_{n,v}$ is below a certain threshold.

The above process is repeated for a number of iterations. In principle frames that are still not annotated (rejected by $g_t$) can be actively labeled by humans, however in practice we have found this situation is rare, unless the distance between the captured views is extremely small, making it difficult to learn a reasonable 3D shape prior.

\vspace{-0.1cm}
\subsection{Outlier detection using multi-view NRSfM network} \label{sec: neural_prior_outlier}
\textbf{Uncertainty score.}
Our geometric constraint function $g$ is built upon measuring the discrepancy between detected 2D landmarks and the 3D reconstruction by a multi-view NRSfM method. This is in the same spirit as using the reprojection error of triangulation to measure uncertainties as in prior works. The idea is if the detected 2D landmarks at different views are all correct, we should be able to recover accurate camera poses and 3D structures, and consequently the reprojection of recovered 3D landmarks matches the 2D landmarks. On the other hand, if the reprojection error is high, it means there exists errors in the 2D landmarks which prevents perfect 3D reconstructions. This leads to the following formulation of our uncertainty score,
\begin{equation}
    y_{(n,v)} = \|\tilde{\mathbf{W}}_{(n,v)}-\text{proj}(\tilde{\mathbf{T}}_{(n,v)}\tilde{\mathbf{S}}_{n})\|_F
    \label{eq:uncertainty_score}
\end{equation}
where $\tilde{\mathbf{T}}_{(n,v)}$, $\tilde{\mathbf{S}}_n$ are the estimated camera extrinsics and 3D landmark positions in the world coordinate, $\tilde{\mathbf{W}}_{(n,v)}$ is 2D landmarks estimated by the landmark detector, and $\text{proj}$ is the projection function. The effectiveness of the uncertainty score defined by Eq.~\ref{eq:uncertainty_score} depends on the reliability of estimating  $\tilde{\mathbf{T}}_{(n,v)}$, $\tilde{\mathbf{S}}_n$. However, due to the low number of synchronized views as well as noise in $\tilde{\mathbf{W}}_{(n,v)}$, simply performing SfM and triangulation gives poor result as shown in Fig.~\ref{Fig: Outlier_rejection_human}. This motivates the following use of MV-NRSfM.

\textbf{Unsupervised learned MV-NRSfM.} Our solution to reliably estimate $\tilde{\mathbf{T}}_{(n,v)}$, $\tilde{\mathbf{S}}_n$ is to marry both the multi-view geometric constraints and the temporal redundancies across frames, which leads to the adaption of the MV-NRSfM method~\cite{dabhi3dv}. Limited by space, we refer interested reader to their paper for detailed treatment. Here we briefly discuss its usage in our problem.
In a nutshell, MV-NRSfM~\cite{dabhi3dv} assumes that 3D shapes (concatenation of 3D landmark positions) can be compressed into low-dimensional latent codes if they are properly aligned to a canonical view. MV-NRSfM is then trained to learn a decoder $h_d: \boldsymbol{\varphi}\in \mathbb{R}^K \rightarrow \mathbf{S}\in\mathbb{R}^{P\times 3}$ which maps a low-dimensional code to an aligned 3D shape, as well as an encoder network $h_e: \mathbf{W}_1, \mathbf{W}_2, ..., \mathbf{W}_V \rightarrow \boldsymbol{\varphi}$ which estimates a single shape code $\boldsymbol{\varphi}$ from 2D landmarks $\mathbf{W}_v\in\mathbb{R}^{P\times 2}$ captured from a number of different views (see Appendix~\ref{appendix: mvnrsfm_arch} for the network architecture). Both $h_d$ and $h_e$ are learned through minimizing the reprojection error:
\begin{equation}
 \min_{\mathbf{T}_{(n,v)}, h_d, h_e} \sum_{(n,v)\in \mathcal{S}}\|\tilde{\mathbf{W}}_{(n,v)} - \text{proj}(\mathbf{T}_{(n,v)} (h_d \circ h_e) (\tilde{\mathbf{W}}_{(n,1)}, \tilde{\mathbf{W}}_{(n,2)}, ..., \tilde{\mathbf{W}}_{(n,V)}))\|_F
\end{equation}
where $\mathcal{S}$ refers to the training set, and $\circ$ denotes function composition. 
Thanks to the constraint from low dimensional codes as well as the convolution structure of $h_e$ inspired from factorization-based NRSfM methods~\cite{deep_nrsfm}, the learned networks $h_d \circ h_e$ are able to infer reasonable 3D landmark positions from noisy 2D landmark inputs. We provide the network architecture of MV-NRSfM in Fig.~\ref{Fig: mv-nrsfm-arch-autoencoder} of Appendix~\ref{appendix: mvnrsfm_arch}.

In our task, we rely on the robustness of MV-NRSfM not only to learn the 3D reconstruction of the labeled training set, but also to detect outliers on the unlabeled set using  Eq.~\ref{eq:uncertainty_score}. At the $t$-th iteration of our self-training, we train $h_d^t$, $h_e^t$ given the current labeled set $\mathcal{S}_{t-1}$ from the previous iteration. We then test $h_d^t \circ h_e^t$ on the detected 2D landmarks from the unlabeled set to produce $\tilde{\mathbf{S}}_n$ used in Eq.~\ref{eq:uncertainty_score}. Camera extrinsics $\tilde{\mathbf{T}}_{(n,v)}$ are then estimated simply through either an orthographic-N-point (OnP) or perspective-N-point (PnP) solver depending on the choice of camera projection model. In our data, we find that assuming a weak perspective camera and use OnP already gives high fidelity results. 

Finally, We note that the unsupervised learned MV-NRSfM networks \ie $h_d \circ h_e$ is likely not able to estimate correct 3D landmarks if its 2D inputs are significantly different than its training set. Instead, it tends to output a plausible 3D structure but does not fully match the 2D inputs. This is actually a desirable behavior for our task, since it serves the purpose of out-of-distribution detection -- detects any shapes differ significantly to the current labeled set. We expect the MV-NRSfM to cover full shape variations in the input sequences as the ``labeled'' set expands while the training progresses. We give a detailed analysis in Fig.~\ref{Fig: 2d_improvements_over_bottleneck} of Appendix~\ref{appendix: iterations}.

\section{Experiments} \label{sec: results}

Our experiments aim to answer the following questions: \textbf{(I)} Is MBW able to generate reliable 2D and 3D landmark predictions from limited views (as few as two) given only a few (as few as 1-2\%) human labels? \textbf{(II)}: Is MBW able to reject outliers and learn a meaningful shape distribution from these few input labels? \textbf{(III)} How important is the number of views in our pipeline? \textbf{(IV)} Can MBW refine (denoise) the 2D candidate inliers? \textbf{(V)} Is our pipeline able to compete with leading benchmarks despite using a fraction of input 2D labels? Before diving into our experiments, we discuss the details of our pipeline.

\vspace{-0.2cm}
\paragraph{Datasets.} Datasets with multi-view videos of non-human subjects are rare, so we collect our own dataset of animals. The collected \textbf{zoo dataset} consists of five animal categories, each with 2 synchronized videos. The videos contain viewpoint and dynamic appearance changes as well as common imaging artifacts such as reflection of water or blurred frames (see Fig.~\ref{fig:dataset_variation}).
For this data we manually annotated part of the sequences for evaluation. In addition, we used the benchmark dataset of Human3.6M~\cite{human36m} (H36M) to perform quantitative evaluations of our approach.

\vspace{-0.2cm}
\paragraph{Implementation Details}
We train our approach on $1$ NVIDIA RTX $3090$ GPU with $24$ GB memory. A learning rate of $0.001$ is used for all networks. We train each network from scratch. A pre-trained RAFT network~\cite{raft} is used with flow iterations of $20$. Bottleneck size of $8$ is used for MV-NRSfM~\cite{dabhi3dv} for all categories. We use HRNet~\cite{hrnet} as the backbone 2D detector, and the same configuration is used for all object categories.

\vspace{-0.2cm}
\paragraph{Question I: Limited amount of labels and views}
We use just two camera views from Directions-1 sequence of Subject \#1 from H36M dataset~\cite{human36m}. Each camera view consists of $1383$ frames per-view, amounting to $2766$ frames in total. Of these, we provide hand labels for only 20 frames (10 frames per view amounting to ~$0.8$\% of the total frames) through uniform sampling. Our task is to generate 2D landmark predictions of the remaining frames of this sequence (~99.2\% unlabeled). 

We evaluate the accuracy of 2D landmark predictions using the commonly used evaluation metric of PCK by Andriluka et al.~\cite{pckh}. We report area under the curve of PCK at different thresholds to understand the nature of 2D prediction errors over all the frames. For consistency, the 2D landmark error is normalized using head bone length before evaluation. As baseline, we keep the complete architecture of MBW, but replace MV-NRSfM with multi-view triangulation using groundtruth calibrated cameras to reject outliers and denoise inliers~\cite{simon_tomas_hand}. We denote this baseline as Trng.

\begin{figure} [!t]
  \centering
   \subfloat[Outlier detection on ~\cite{human36m}. ]{\vspace{-3cm}{\includegraphics[width=0.32\linewidth]{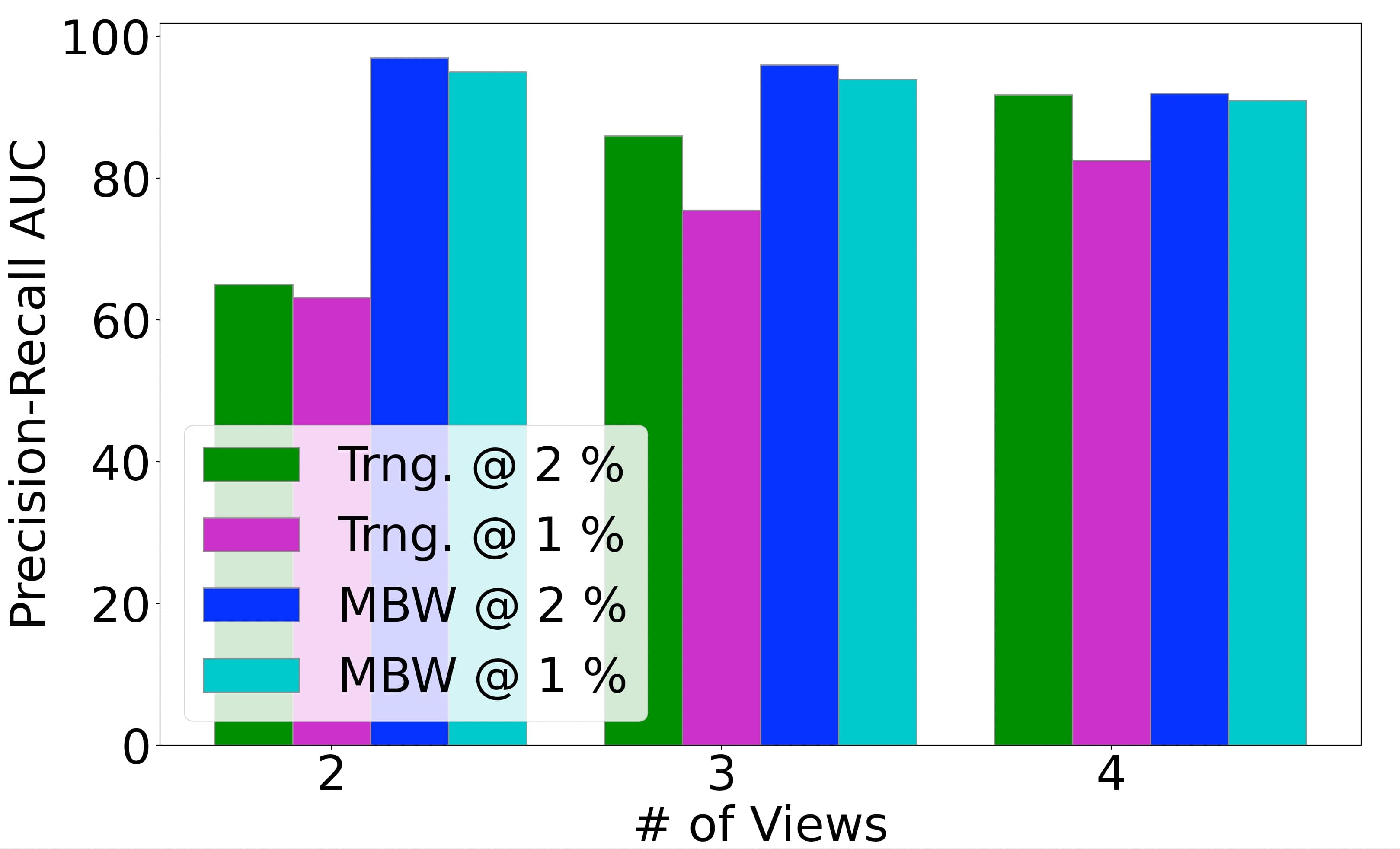} }\label{Fig: Outlier_rejection_human}}  
  \subfloat[Outlier detection on Zoo dataset ]{\vspace{-3cm}{\includegraphics[width=0.32\linewidth]{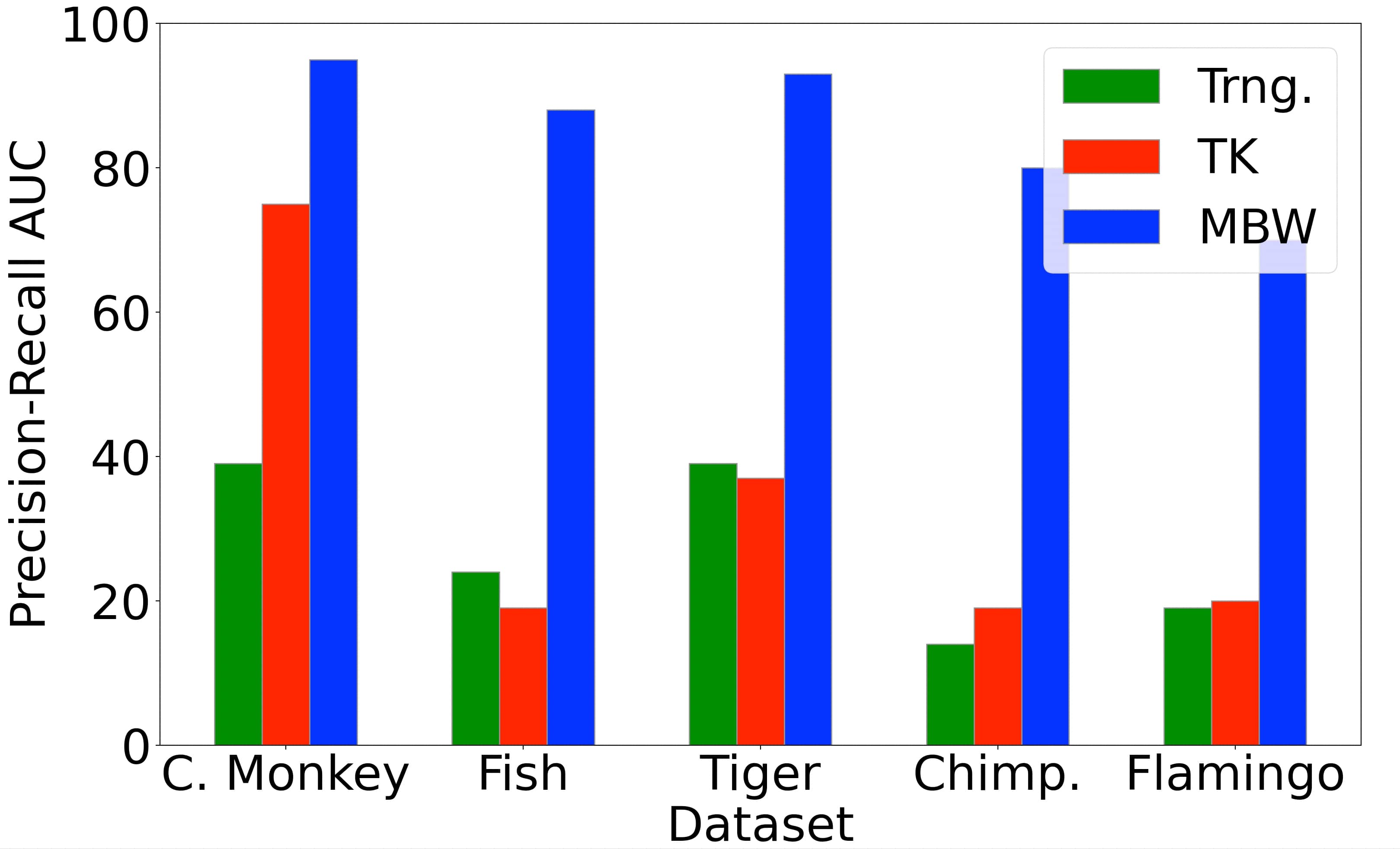} }\label{Fig: Outlier_rejection_Zoo}}
  \subfloat[2D accuracy using PCKh~\cite{pckh}. ]{\vspace{-3cm}{\includegraphics[width=0.32\linewidth]{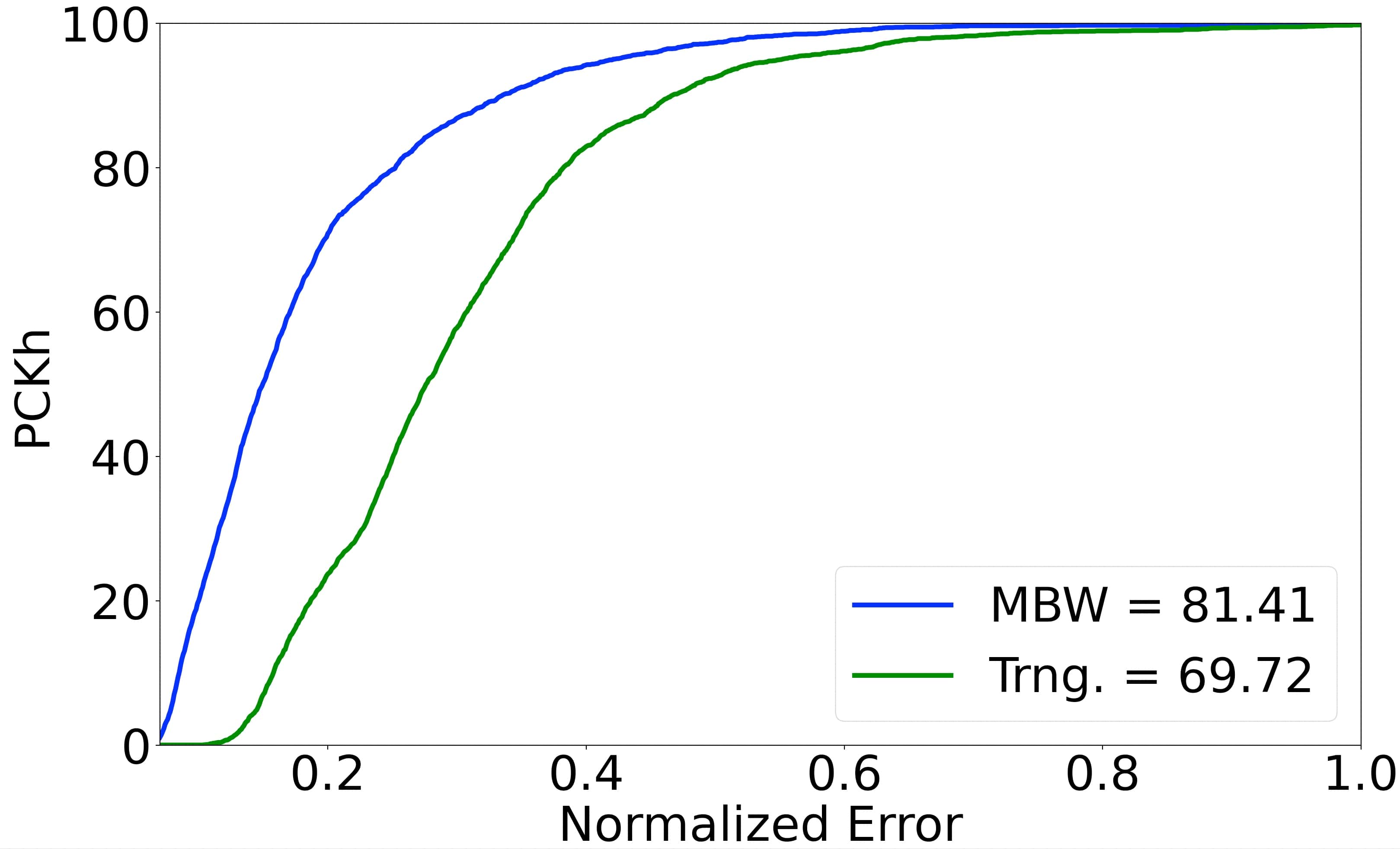} }\label{Fig: 2d_training_1}}
  \caption{(a) Precision-Recall (PR) AUC shows the outlier rejection accuracy on benchmark~\cite{human36m} with varying number of camera views and input 2D labels. PR AUC bars per each view are calculated using the frames available in that view. Fewer views corresponds to a smaller number of frames, and more views comprise a larger number of frames. (b) Outlier rejection accuracy using PR AUC on zoo dataset. (c) Final 2D landmark prediction accuracy using our approach compared to multi-view triangulation (with groundtruth cameras)~\cite{hartley1997triangulation} (Trng). We calculate the normalized error (dividing by head bone length~\cite{pckh}) and report the PCK AUC result varying the error threshold range. Note that we run this experiment on just $10$ frames per view (amounting to~$0.8\%$) and only two views. }
  \label{Fig: Outlier_rejection}
  \vspace{-0.2cm}
\end{figure}

2D landmark prediction performance over all the frames is shown in Fig.~\ref{Fig: 2d_training_1}. The quantitative results are shown in Tab.~\ref{Tab: 3D_2D_0_2} where we we report PCK AUC values to evaluate 2D landmark prediction accuracy. We evaluate the 3D structure accuracy using Procrustes-Aligned Mean Per Joint Position Error (PA-MPJPE)~\cite{mpjpe}. This metric evaluates 3D joint localization accuracy in mm and represents the $L2$ distance between the groundtruth and predicted joint locations after aligning the 3D structures using a rigid transformation. Table~\ref{Tab: 3D_2D_0_2} shows that our approach is able to generate high-fidelity 2D landmark prediction as well as accurate 3D structure despite starting from a mere $0.8 \%$ of 2-View data. In contrast, the competing baseline fails since it cannot reconstruct good 3D structure from just $2$ views and extremely noisy landmark predictions. This experiment helps us answer Question \textbf{(1)}: {\bf Yes}, MBW with MV-NRSfM is able to predict reasonable 2D and 3D landmark prediction using small amount of labels and views.

\begin{table} 
\caption{Quantitative comparison showing improvement in 2D detection compared to comparative baselines (with groundtruth camera) on benchmark dataset~\cite{human36m}.}
  \label{Tab: 3D_2D_0_2}
  \centering
  \begin{tabular}{ccc}
    \toprule
    Method    & 2D $\boldsymbol{\uparrow}$ & 3D (mm) $\boldsymbol{\downarrow}$ \\
    \midrule
    Triangulation~\cite{hartley1997triangulation} & $69.72$ & $1383$   \\    
    MBW (Ours) & $\mathbf{81.41}$ & $\mathbf{39.3}$ \\
    \bottomrule
  \end{tabular}
  \vspace{-0.5cm}
\end{table}

\vspace{-0.2cm}
\paragraph{Question II: Outlier rejection} The proposed pipeline requires bad 2D landmark candidates (outliers) to be rejected so they are not incorporated into subsequent training iterations as (inlier) pseudo-labels ${\mathcal{S}_{i}}$. To evaluate this, we design the following experiment to evaluate the Precision-Recall AUC on our \textbf{zoo} dataset as shown in Fig.~\ref{Fig: Outlier_rejection_Zoo}. The number of initial 2D manual labels provided to each object is $~2\%$. 
We perform an ablation study in which we evaluate at the initial iteration of MBW, to check MBW's ability to reject and refine noisy landmark predictions from optical flow.

MBW is compared against weak-perspective Tomasi-Kanade Structure-from-Motion (TK)~\cite{Tomasi92shapeand} -- the only other approach able to reconstruct 3D structure without calibrated cameras. For completeness, we also estimate cameras in the wild using ~\cite{hloc} so that we can compare against Trng~\cite{hartley_multiview_geometry}. For the competing approaches, we use our whole pipeline as-is except for the MV-NRSfM outlier rejection and denoising part, instead using the competing method. Since there is no groundtruth in the \textbf{zoo} dataset, we manually label each 2D flow prediction and assign inlier labels by visual verification. The PR AUC is calculated using this manually generated groundtruth verification. As evident in Fig.~\ref{Fig: Outlier_rejection_Zoo}, only our approach can accurately reject the outliers as evident by high PR AUC. This finding is reasonable since we found that~\cite{hloc} is unable to calculate cameras due to non-existent or poor matching features across views collected in the \textbf{zoo} dataset, resulting in no 3D reconstruction. Another baseline, weak-perspective Tomasi Kanade SfM is unable to handle the noise encountered in real-world data and hence fails to accurately reconstruct 3D. Thus, the above experiment helps us answer Question \textbf{(II)}: {\bf Yes}, MBW is able to reject outliers despite of having limited input 2D labels, number of views, and no camera information.

\vspace{-0.2cm}
\paragraph{Question III: Number of Views} For this experiment, we take the same H36M dataset as noted in Question 1, with the exception that we vary the number of views (2-4) and number of 2D input labels (1-2\%) provided to our approach and competing baseline of Trng. In Fig.~\ref{Fig: Outlier_rejection_human} we see that both our approach and the baseline reject outliers reasonably well for four views. However as we reduce the number of views Trng. had a steeper dropoff due to unreliable geometric constraints arising from limited views and noisy 2D labels. In contrast, since MBW uses a learned shape prior for rejecting outliers, our performance remains consistent across varying views. This experiment helps us answer Question \textbf{(III)}: {\bf Yes}, as long as we learn a good shape prior, we do not require large number of views to reliably reject the outliers.

\begin{figure} [!t]
  \centering
  {\includegraphics[width=0.95\linewidth]{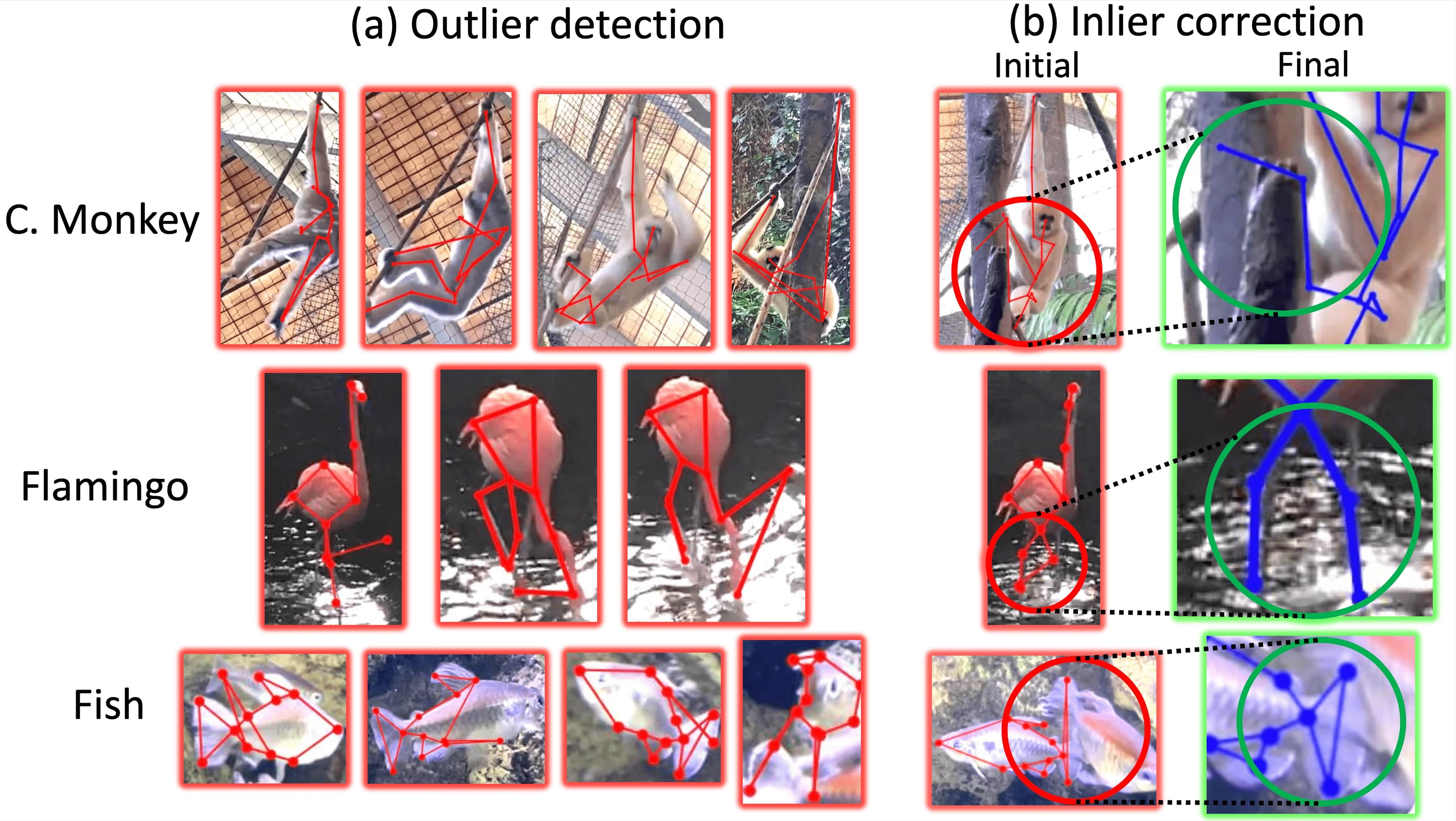}}
  \caption{(a) Shows example outliers detected using the proposed approach. (b) Shows the denoising capability of the proposed approach where we are able to denoise the improve the inliers by using the multi-view shape prior information.}
  \label{Fig: Visualization_Improvements}
  \vspace{-0.3cm}
\end{figure} 
\textbf{Question IV: Denoising capability} To answer this question, we report the visual results of the 2D landmark predictions by MBW on \textbf{zoo} dataset in Fig.~\ref{Fig: Visualization_Improvements}. On the left, we see that MBW is able to detect bad 2D predictions without human intervention (also evident by PR AUC in Fig.~\ref{Fig: Outlier_rejection_Zoo}). On the right, we highlight scenarios where our approach can use the MV-NRSfM shape prior information to not only detect but refine (denoise) the inliers. Moreover, apart from qualitative visual inspection, we also conduct a toy study to analyze the denoising capabilities of our approach and discuss this in the supplementary section. The visual inspection experiment in Fig.~\ref{Fig: Visualization_Improvements} helps us answer Question \textbf{(IV)}: {\bf Yes}, MBW can effectively denoise candidate inliers.

\begin{table}
  \caption{3D pose reconstruction accuracy of different methods on the Human3.6M dataset~\cite{human36m}. Our approach uses just $2\%$ of 2D labels yet achieves 3D performance comparable to competing methods.}
  \label{Tab: 3D_SOTA}
  \centering
  \begin{tabular}{ll}
    \toprule
    Multi-view methods & (mm) $\boldsymbol{\downarrow}$ \\
    \midrule
    Martinez et al. (multi-view).~\cite{martinez}     &  $46.5$ \\
    Pavlakos et al. ~\cite{Pavlakos} & $41.2$ \\
    Kadkhodamohammadi \& Padoy ~\cite{kadkhodamohammadi2021generalizable}  &  $39.4$\\
    Iskakov et al.~\cite{learnableTrng} & $19.9$\\
    Reddy et al.~\cite{tessetrack}     &  $\mathbf{17.5}$ \\
    \midrule
    Ours (PA-MPJPE at 2\%)  & $20.15$\\
    \bottomrule
  \end{tabular}
  \vspace{-0.3cm}
\end{table}
\begin{table}
  \caption{We compare our approach with existing semi-supervised learning frameworks: (1) temporal supervision~\cite{sbr} and (2) cross-view supervision~\cite{monet,msbr}. We evaluate on human dataset (Human3.6M~\cite{human36m}) using PCKh measure. We test the generalizability by applying on unseen data. We use just $\mathbf{2}\%$ of the labeled data compared to other approaches.}
  \label{Tab: 3D_2D_acc}
  \centering
  \begin{tabular}{llllllll}
    \toprule
    Human3.6M~\cite{human36m}     & Nec $\boldsymbol{\uparrow}$ & Sho $\boldsymbol{\uparrow}$ & Elb $\boldsymbol{\uparrow}$ & Wri $\boldsymbol{\uparrow}$ & Hip $\boldsymbol{\uparrow}$ & Kne $\boldsymbol{\uparrow}$ & Ank $\boldsymbol{\uparrow}$ \\
    \midrule
    Dong et al.~\cite{sbr} & $91.7$ & $81.4$ & $42.3$ & $25.6$ & $93.9$ & $83.4$ & $87.5$  \\
    Jafarian et al.~\cite{monet} & $89.6$ & $48.3$ & $29.7$ & $20.5$ & $29.8$ & $34.9$ & $60.7$    \\
    Zheng \& Park~\cite{msbr} & $93.2$ & $\mathbf{92.8}$ & $67.3$  & $49.6$ & $\mathbf{93.7}$ & $\mathbf{87.6}$ & $89.5$         \\    
    \midrule
    MBW (Ours) & $\mathbf{96.8}$ & ${83.3}$ & $\mathbf{78.1}$ & $\mathbf{69.8}$  & ${89.2}$ & ${82.9}$ & $\mathbf{92.9}$       \\    
    \bottomrule
  \end{tabular}
  \vspace{-0.3cm}
\end{table}

\vspace{-0.2cm}
\paragraph{Question V: Comparison against leading benchmarks} Finally, we compare the performance of our approach with existing state-of-the-art semi-supervised learning approaches. Since our approach has overlap with MSBR~\cite{msbr} in that they have an initial labeled set, as well as an unlabeled set, we compare and report the 2D landmark prediction accuracy using the same experimental setup as theirs and report in Tab.~\ref{Tab: 3D_2D_acc}. To reiterate, similar to them we use Eating and Discussion but just use 2\% as input labeled data and test on all frames of Greeting for a fair comparison. We observe that in spite of using a small amount of 2D input labels, our approach is able to compete against well-established benchmark approaches that use $>$30\% of frames to label -- making our approach more efficient in terms of effort required to generate labels of similar accuracy.

Lastly, we conduct 3D reconstruction analysis on the Directions-1 sequence of H36M~\cite{human36m} using the common protocol where Subjects 1,5,6,7,8 are used as train sets, and S9, and 11 are used as the test set. We just provide 2\% of train set labels as 2D input labels to our approach and evaluate the 3D reconstruction performance on 100\% of the test set, and report the 3D reconstruction results in Tab.~\ref{Tab: 3D_SOTA}. Since our approach calculates 3D reconstruction in a canonical frame and is up-to-scale, we apply the rigid transformation of Procrustes Alignment and report the 3D landmark reconstruction accuracy as PA-MPJPE in mm. We observe that we are able to compete with state-of-the-art benchmarks despite using 2\% input labels and no camera information in our approach. The above two experiments helps us answer Question \textbf{(V)}.

\section{Limitations} \label{sec: limitation}
\vspace{-0.2cm}
Our approach relies on a 3D neural shape prior to address challenging scenarios such as occluded or noisy keypoints. However, it is only able make accurate predictions if the views have sufficiently wide baseline, for instance to address cases where an occluded keypoint in one view is unoccluded in another view. In these scenarios  MBW can reject (or clean) the occluded keypoint through the 3D prior. However if the keypoint is occluded in both (all) the available views, then MBW fails to produce meaningful predictions. This is expected since MBW cannot hallucinate without any cues or priors. Some more recent works such as particle trajectory tracking~\cite{harley2022particle} may help address the issues of occlusion, and we plan to take these lines of work into account in future work.

\section{Conclusion}
\vspace{-0.2cm}
The idea of leveraging geometry to make labeling easier is not new -- indeed this is the basis for large-scale dome-based data collection methods~\cite{panoptic,openmonkeystudio,InterHand,contrahand,dogs_dataset}. Our key contribution is incorporating a neural prior to enable the same ideas to be applied to tail-end distribution non-rigid object categories. This has yielded several key insights: \textbf{(a)} Deep 3D neural shape priors trained on even as few as $10-15$ frames per video can already provide powerful constraints for modeling deformable objects; \textbf{(b)} provided a reliable outlier rejection method, optical flow-based methods~\cite{raft} provide a simple and effective way to propagate 2D candidate predictions in videos; 
\textbf{(c)} neural priors can be used as the basis of reliable outlier rejection, enabling bootstrapping under constrained and error-prone methods such as flow and ``partially trained'' 2D networks into highly accurate vision pipelines. Our approach is just a first step in this direction, but it may point the way to a revival of iterative pipelines that more deeply integrate the insights of recent neural approaches with iterative refinement methods familiar in classic 3D computer vision.

{\small
\bibliographystyle{plainnat}
\bibliography{references}

\begin{thebibliography}{44}
\providecommand{\natexlab}[1]{#1}
\providecommand{\url}[1]{\texttt{#1}}
\expandafter\ifx\csname urlstyle\endcsname\relax
  \providecommand{\doi}[1]{doi: #1}\else
  \providecommand{\doi}{doi: \begingroup \urlstyle{rm}\Url}\fi

\bibitem[Bala et~al.(2020)Bala, Eisenreich, Yoo, Hayden, Park, and
  Zimmermann]{openmonkeystudio}
Praneet~C Bala, Benjamin~R Eisenreich, Seng Bum~Michael Yoo, Benjamin~Y Hayden,
  Hyun~Soo Park, and Jan Zimmermann.
\newblock Openmonkeystudio: Automated markerless pose estimation in freely
  moving macaques.
\newblock \emph{BioRxiv}, 2020.

\bibitem[Dabhi(2022)]{Dabhi_MBW_Zoo_data}
Mosam Dabhi.
\newblock {MBW Zoo data}, 2022.
\newblock URL \url{https://github.com/mosamdabhi/MBW-Data}.

\bibitem[Dabhi et~al.(2021)Dabhi, Wang, Saluja, Jeni, Fasel, and
  Lucey]{dabhi3dv}
Mosam Dabhi, Chaoyang Wang, Kunal Saluja, L{\'a}szl{\'o}~A Jeni, Ian Fasel, and
  Simon Lucey.
\newblock High fidelity 3d reconstructions with limited physical views.
\newblock In \emph{2021 International Conference on 3D Vision (3DV)}, pages
  1301--1311. IEEE, 2021.

\bibitem[Dong et~al.(2020)Dong, Yang, Wei, Weng, Sheikh, and Yu]{sbr}
Xuanyi Dong, Yi~Yang, Shih-En Wei, Xinshuo Weng, Yaser Sheikh, and Shoou-I Yu.
\newblock Supervision by registration and triangulation for landmark detection.
\newblock \emph{IEEE transactions on pattern analysis and machine
  intelligence}, 43\penalty0 (10):\penalty0 3681--3694, 2020.

\bibitem[Felzenszwalb and Huttenlocher(2005)]{pictorial_structure}
Pedro~F Felzenszwalb and Daniel~P Huttenlocher.
\newblock Pictorial structures for object recognition.
\newblock \emph{International journal of computer vision}, 61\penalty0
  (1):\penalty0 55--79, 2005.

\bibitem[Feng et~al.(2021)Feng, He, Wen, Keskin, and Ye]{active_learning_meta}
Qi~Feng, Kun He, He~Wen, Cem Keskin, and Yuting Ye.
\newblock Active learning with pseudo-labels for multi-view 3d pose estimation.
\newblock \emph{arXiv preprint arXiv:2112.13709}, 2021.

\bibitem[Gebru et~al.(2021)Gebru, Morgenstern, Vecchione, Vaughan, Wallach,
  Iii, and Crawford]{datasheets}
Timnit Gebru, Jamie Morgenstern, Briana Vecchione, Jennifer~Wortman Vaughan,
  Hanna Wallach, Hal~Daum{\'e} Iii, and Kate Crawford.
\newblock Datasheets for datasets.
\newblock \emph{Communications of the ACM}, 64\penalty0 (12):\penalty0 86--92,
  2021.

\bibitem[G{\"u}nel et~al.(2019)G{\"u}nel, Rhodin, Morales, Campagnolo, Ramdya,
  and Fua]{deepfly3d}
Semih G{\"u}nel, Helge Rhodin, Daniel Morales, Jo{\~a}o Campagnolo, Pavan
  Ramdya, and Pascal Fua.
\newblock Deepfly3d, a deep learning-based approach for 3d limb and appendage
  tracking in tethered, adult drosophila.
\newblock \emph{Elife}, 8:\penalty0 e48571, 2019.

\bibitem[Harley et~al.(2022)Harley, Fang, and Fragkiadaki]{harley2022particle}
Adam~W Harley, Zhaoyuan Fang, and Katerina Fragkiadaki.
\newblock Particle videos revisited: Tracking through occlusions using point
  trajectories.
\newblock \emph{arXiv preprint arXiv:2204.04153}, 2022.

\bibitem[Hartley and Zisserman(2003)]{hartley_multiview_geometry}
Richard Hartley and Andrew Zisserman.
\newblock \emph{Multiple view geometry in computer vision}.
\newblock Cambridge university press, 2003.

\bibitem[Hartley and Sturm(1997)]{hartley1997triangulation}
Richard~I Hartley and Peter Sturm.
\newblock Triangulation.
\newblock \emph{Computer vision and image understanding}, 68\penalty0
  (2):\penalty0 146--157, 1997.

\bibitem[Ionescu et~al.(2013)Ionescu, Papava, Olaru, and
  Sminchisescu]{human36m}
Catalin Ionescu, Dragos Papava, Vlad Olaru, and Cristian Sminchisescu.
\newblock Human3.6m: Large scale datasets and predictive methods for 3d human
  sensing in natural environments.
\newblock \emph{IEEE transactions on pattern analysis and machine
  intelligence}, 36\penalty0 (7):\penalty0 1325--1339, 2013.

\bibitem[Iskakov et~al.(2019)Iskakov, Burkov, Lempitsky, and
  Malkov]{learnableTrng}
Karim Iskakov, Egor Burkov, Victor Lempitsky, and Yury Malkov.
\newblock Learnable triangulation of human pose.
\newblock In \emph{Proceedings of the IEEE/CVF International Conference on
  Computer Vision}, pages 7718--7727, 2019.

\bibitem[Joo et~al.(2015)Joo, Liu, Tan, Gui, Nabbe, Matthews, Kanade, Nobuhara,
  and Sheikh]{panoptic}
Hanbyul Joo, Hao Liu, Lei Tan, Lin Gui, Bart Nabbe, Iain Matthews, Takeo
  Kanade, Shohei Nobuhara, and Yaser Sheikh.
\newblock Panoptic studio: A massively multiview system for social motion
  capture.
\newblock In \emph{Proceedings of the IEEE International Conference on Computer
  Vision}, pages 3334--3342, 2015.

\bibitem[Joska et~al.(2021)Joska, Clark, Muramatsu, Jericevich, Nicolls,
  Mathis, Mathis, and Patel]{acinoset}
Daniel Joska, Liam Clark, Naoya Muramatsu, Ricardo Jericevich, Fred Nicolls,
  Alexander Mathis, Mackenzie~W Mathis, and Amir Patel.
\newblock Acinoset: A 3d pose estimation dataset and baseline models for
  cheetahs in the wild.
\newblock In \emph{2021 IEEE International Conference on Robotics and
  Automation (ICRA)}, pages 13901--13908. IEEE, 2021.

\bibitem[Kadkhodamohammadi and Padoy(2021)]{kadkhodamohammadi2021generalizable}
Abdolrahim Kadkhodamohammadi and Nicolas Padoy.
\newblock A generalizable approach for multi-view 3d human pose regression.
\newblock \emph{Machine Vision and Applications}, 32\penalty0 (1):\penalty0
  1--14, 2021.

\bibitem[Kearney et~al.(2020)Kearney, Li, Parsons, Kim, and
  Cosker]{dogs_dataset}
Sinead Kearney, Wenbin Li, Martin Parsons, Kwang~In Kim, and Darren Cosker.
\newblock Rgbd-dog: Predicting canine pose from rgbd sensors.
\newblock In \emph{Proceedings of the IEEE/CVF Conference on Computer Vision
  and Pattern Recognition}, pages 8336--8345, 2020.

\bibitem[Kong and Lucey(2019)]{deep_nrsfm}
Chen Kong and Simon Lucey.
\newblock Deep non-rigid structure from motion.
\newblock In \emph{Proceedings of the IEEE International Conference on Computer
  Vision}, pages 1558--1567, 2019.

\bibitem[Kumar et~al.(2020)Kumar, Van~Gool, de~Oliveira, Cherian, Dai, and
  Li]{dense_nrsfm}
Suryansh Kumar, Luc Van~Gool, Carlos~EP de~Oliveira, Anoop Cherian, Yuchao Dai,
  and Hongdong Li.
\newblock Dense non-rigid structure from motion: A manifold viewpoint.
\newblock \emph{arXiv preprint arXiv:2006.09197}, 2020.

\bibitem[Marshall et~al.(2021)Marshall, Klibaite, gellis, Aldarondo, Olveczky,
  and Dunn]{rats-large-scale}
Jesse Marshall, Ugne Klibaite, amanda gellis, Diego Aldarondo, Bence Olveczky,
  and Timothy~W Dunn.
\newblock The pair-r24m dataset for multi-animal 3d pose estimation.
\newblock In J.~Vanschoren and S.~Yeung, editors, \emph{Proceedings of the
  Neural Information Processing Systems Track on Datasets and Benchmarks},
  volume~1, 2021.

\bibitem[Martinez et~al.(2017)Martinez, Hossain, Romero, and Little]{martinez}
Julieta Martinez, Rayat Hossain, Javier Romero, and James~J Little.
\newblock A simple yet effective baseline for 3d human pose estimation.
\newblock In \emph{Proceedings of the IEEE International Conference on Computer
  Vision}, pages 2640--2649, 2017.

\bibitem[Mathis et~al.(2018)Mathis, Mamidanna, Cury, Abe, Murthy, Mathis, and
  Bethge]{deeplabcut}
Alexander Mathis, Pranav Mamidanna, Kevin~M Cury, Taiga Abe, Venkatesh~N
  Murthy, Mackenzie~Weygandt Mathis, and Matthias Bethge.
\newblock Deeplabcut: markerless pose estimation of user-defined body parts
  with deep learning.
\newblock \emph{Nature neuroscience}, 21\penalty0 (9):\penalty0 1281--1289,
  2018.

\bibitem[Mildenhall et~al.(2020)Mildenhall, Srinivasan, Tancik, Barron,
  Ramamoorthi, and Ng]{nerf}
Ben Mildenhall, Pratul~P Srinivasan, Matthew Tancik, Jonathan~T Barron, Ravi
  Ramamoorthi, and Ren Ng.
\newblock Nerf: Representing scenes as neural radiance fields for view
  synthesis.
\newblock In \emph{European conference on computer vision}, pages 405--421.
  Springer, 2020.

\bibitem[Moon et~al.(2020)Moon, Yu, Wen, Shiratori, and Lee]{InterHand}
Gyeongsik Moon, Shoou-I Yu, He~Wen, Takaaki Shiratori, and Kyoung~Mu Lee.
\newblock Interhand2.6m: A dataset and baseline for 3d interacting hand pose
  estimation from a single rgb image.
\newblock In \emph{European Conference on Computer Vision (ECCV)}, 2020.

\bibitem[Newell et~al.(2016)Newell, Yang, and Deng]{stacked_hourglass}
Alejandro Newell, Kaiyu Yang, and Jia Deng.
\newblock Stacked hourglass networks for human pose estimation.
\newblock In \emph{European conference on computer vision}, pages 483--499.
  Springer, 2016.

\bibitem[Pavlakos et~al.(2017)Pavlakos, Zhou, Derpanis, and
  Daniilidis]{Pavlakos}
Georgios Pavlakos, Xiaowei Zhou, Konstantinos~G Derpanis, and Kostas
  Daniilidis.
\newblock Harvesting multiple views for marker-less 3d human pose annotations.
\newblock In \emph{Proceedings of the IEEE conference on computer vision and
  pattern recognition}, pages 6988--6997, 2017.

\bibitem[Pereira et~al.(2022)Pereira, Tabris, Matsliah, Turner, Li,
  Ravindranath, Papadoyannis, Normand, Deutsch, Wang, et~al.]{sleap}
Talmo~D Pereira, Nathaniel Tabris, Arie Matsliah, David~M Turner, Junyu Li,
  Shruthi Ravindranath, Eleni~S Papadoyannis, Edna Normand, David~S Deutsch,
  Z~Yan Wang, et~al.
\newblock Sleap: A deep learning system for multi-animal pose tracking.
\newblock \emph{Nature methods}, pages 1--10, 2022.

\bibitem[Reddy et~al.(2021)Reddy, Guigues, Pishchulin, Eledath, and
  Narasimhan]{tessetrack}
N~Dinesh Reddy, Laurent Guigues, Leonid Pishchulin, Jayan Eledath, and
  Srinivasa~G Narasimhan.
\newblock Tessetrack: End-to-end learnable multi-person articulated 3d pose
  tracking.
\newblock In \emph{Proceedings of the IEEE/CVF Conference on Computer Vision
  and Pattern Recognition}, pages 15190--15200, 2021.

\bibitem[Rhodin et~al.(2018)Rhodin, Sp{\"o}rri, Katircioglu, Constantin, Meyer,
  M{\"u}ller, Salzmann, and Fua]{humanskiing}
Helge Rhodin, J{\"o}rg Sp{\"o}rri, Isinsu Katircioglu, Victor Constantin,
  Fr{\'e}d{\'e}ric Meyer, Erich M{\"u}ller, Mathieu Salzmann, and Pascal Fua.
\newblock Learning monocular 3d human pose estimation from multi-view images.
\newblock In \emph{Proceedings of the IEEE Conference on Computer Vision and
  Pattern Recognition}, pages 8437--8446, 2018.

\bibitem[Sand and Teller(2008)]{sand2008particle}
Peter Sand and Seth Teller.
\newblock Particle video: Long-range motion estimation using point
  trajectories.
\newblock \emph{International Journal of Computer Vision}, 80\penalty0
  (1):\penalty0 72--91, 2008.

\bibitem[Sarlin et~al.(2019)Sarlin, Cadena, Siegwart, and Dymczyk]{hloc}
Paul-Edouard Sarlin, Cesar Cadena, Roland Siegwart, and Marcin Dymczyk.
\newblock From coarse to fine: Robust hierarchical localization at large scale.
\newblock In \emph{CVPR}, 2019.

\bibitem[Sigal et~al.(2010)Sigal, Balan, and Black]{mpjpe}
Leonid Sigal, Alexandru~O Balan, and Michael~J Black.
\newblock Humaneva: Synchronized video and motion capture dataset and baseline
  algorithm for evaluation of articulated human motion.
\newblock \emph{International journal of computer vision}, 87\penalty0
  (1):\penalty0 4--27, 2010.

\bibitem[Simon et~al.(2017)Simon, Joo, Matthews, and Sheikh]{simon_tomas_hand}
Tomas Simon, Hanbyul Joo, Iain Matthews, and Yaser Sheikh.
\newblock Hand keypoint detection in single images using multiview
  bootstrapping.
\newblock In \emph{Proceedings of the IEEE conference on Computer Vision and
  Pattern Recognition}, pages 1145--1153, 2017.

\bibitem[Sun et~al.(2019)Sun, Xiao, Liu, and Wang]{hrnet}
Ke~Sun, Bin Xiao, Dong Liu, and Jingdong Wang.
\newblock Deep high-resolution representation learning for human pose
  estimation.
\newblock In \emph{Proceedings of the IEEE conference on computer vision and
  pattern recognition}, pages 5693--5703, 2019.

\bibitem[Teed and Deng(2020)]{raft}
Zachary Teed and Jia Deng.
\newblock Raft: Recurrent all-pairs field transforms for optical flow.
\newblock In \emph{European conference on computer vision}, pages 402--419.
  Springer, 2020.

\bibitem[Tomasi and Kanade(1992)]{Tomasi92shapeand}
Carlo Tomasi and Takeo Kanade.
\newblock Shape and motion from image streams under orthography: a
  factorization method.
\newblock \emph{International Journal of Computer Vision}, 9\penalty0
  (2):\penalty0 137--154, 1992.

\bibitem[Wei et~al.(2016)Wei, Ramakrishna, Kanade, and Sheikh]{cpm}
Shih-En Wei, Varun Ramakrishna, Takeo Kanade, and Yaser Sheikh.
\newblock Convolutional pose machines.
\newblock In \emph{Proceedings of the IEEE conference on Computer Vision and
  Pattern Recognition}, pages 4724--4732, 2016.

\bibitem[Yang and Ramanan(2012)]{pckh}
Yi~Yang and Deva Ramanan.
\newblock Articulated human detection with flexible mixtures of parts.
\newblock \emph{IEEE transactions on pattern analysis and machine
  intelligence}, 35\penalty0 (12):\penalty0 2878--2890, 2012.

\bibitem[Yao et~al.(2019)Yao, Jafarian, and Park]{monet}
Yuan Yao, Yasamin Jafarian, and Hyun~Soo Park.
\newblock Monet: Multiview semi-supervised keypoint detection via epipolar
  divergence.
\newblock In \emph{Proceedings of the IEEE/CVF International Conference on
  Computer Vision}, pages 753--762, 2019.

\bibitem[Yu et~al.(2021)Yu, Xu, Zhang, Zhao, Guan, and Tao]{ap10k}
Hang Yu, Yufei Xu, Jing Zhang, Wei Zhao, Ziyu Guan, and Dacheng Tao.
\newblock Ap-10k: A benchmark for animal pose estimation in the wild.
\newblock In \emph{Thirty-fifth Conference on Neural Information Processing
  Systems Datasets and Benchmarks Track (Round 2)}, 2021.

\bibitem[Zhang et~al.(2017)Zhang, Liu, Zhou, Leung, and Chan]{humanmartial}
Weichen Zhang, Zhiguang Liu, Liuyang Zhou, Howard Leung, and Antoni~B Chan.
\newblock Martial arts, dancing and sports dataset: A challenging stereo and
  multi-view dataset for 3d human pose estimation.
\newblock \emph{Image and Vision Computing}, 61:\penalty0 22--39, 2017.

\bibitem[Zhang and Park(2020)]{msbr}
Yilun Zhang and Hyun~Soo Park.
\newblock Multiview supervision by registration.
\newblock In \emph{Proceedings of the IEEE/CVF Winter Conference on
  Applications of Computer Vision}, pages 420--428, 2020.

\bibitem[Zimmermann et~al.(2019)Zimmermann, Ceylan, Yang, Russell, Argus, and
  Brox]{freihand}
Christian Zimmermann, Duygu Ceylan, Jimei Yang, Bryan Russell, Max Argus, and
  Thomas Brox.
\newblock Freihand: A dataset for markerless capture of hand pose and shape
  from single rgb images.
\newblock In \emph{Proceedings of the IEEE International Conference on Computer
  Vision}, pages 813--822, 2019.

\bibitem[Zimmermann et~al.(2021)Zimmermann, Argus, and Brox]{contrahand}
Christian Zimmermann, Max Argus, and Thomas Brox.
\newblock Contrastive representation learning for hand shape estimation.
\newblock In \emph{DAGM German Conference on Pattern Recognition}, pages
  250--264. Springer, 2021.

\end{thebibliography}
}


\pagebreak
\begin{center}
\textbf{\Large Supplementary Material}
\end{center}

\appendix

\section{Ablation study - Iterations in MBW} \label{appendix: iterations}

In this section, we conduct an ablation study analyzing the effects of iterations in our proposed approach. In other words, we discuss the improvements in 2D and 3D landmark predictions as well as the implications of iterations in our proposed approach. We conduct our ablation study on the publicly available human benchmark dataset~\cite{human36m}. We initialise our pipeline (MBW) with $5\%$ 2D input labels and $4$ views from ``Directions-1'' sequence of ``Subject-1''~\cite{human36m}. 

\textbf{Improvement in 2D and 3D landmark predictions}: As shown in Fig.~\ref{Fig: 2d_error}, we see that the major improvement in 2D landmark predictions could be observed between Iteration 0 and Iteration 1. Moreover, we see that as the iterations progress, the 2D landmark prediction error continues to reduce as seen in Fig.~\ref{Fig: 2d_error}. Furthermore, we notice that as the iterations progress, our pipeline continues to further denoise and improve the 2D landmark predictions as well as continues to generate accurate pseudo-labels as visible in Fig.~\ref{Fig: occlusion_stage_3_improvements}. Similarly, we see that as the iterations progress, the 3D reconstruction error (in PA-MPJPE) continues to decrease as visible in Tab.~\ref{Tab: 3D_over_iterations}.

\begin{figure} [!t]
    \centering
    \includegraphics[width=0.9\linewidth]{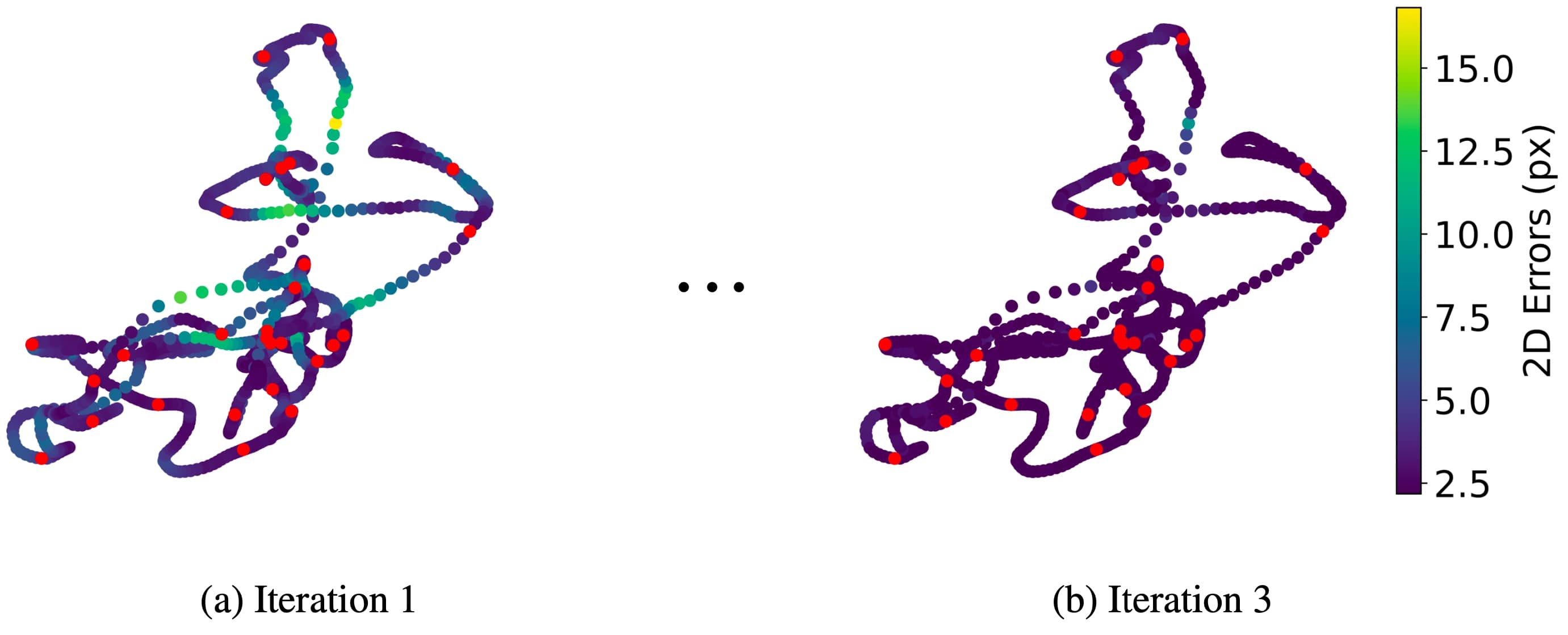}
  \caption{We plot the first two dimensions of the MV-NRSfM bottleneck since it shows the overall coverage and distribution of the sequential data. The red points represent the frames that were given initial 2D input labels. The colorbar of these scatter plots represents the reprojection error (Eq.~\eqref{eq:uncertainty_score}) between the projected 3D structure from MV-NRSfM and 2D candidate predictions at the corresponding iteration. We observe that during the final iteration, the error reduces substantially as visible by colorbars indicating that the MV-NRSfM network generalizes reasonably well with limited initial labels, and during iterations, it expands its coverage over most of the samples in the dataset.}
  \label{Fig: 2d_improvements_over_bottleneck}
  \vspace{-0.1cm}
\end{figure}

We also graphically visualize the effects of MBW at each stage in Fig.~\ref{Fig: 2d_improvements_over_bottleneck}. With a learned MV-NRSfM over given data, we visualize the first two dimensions of the bottleneck. The initial two dimensions of the bottleneck show the overall spread of the given data. The red dots in this plot represents the initial set of 2D input labels. We color code this scatter plot based on 2D reprojection error. Specifically, the colors in Fig.~\ref{Fig: 2d_improvements_over_bottleneck} represent the error calculated from Eq.~\eqref{eq:uncertainty_score}. As the iterations progress, we observe the reprojection error to continue to decrease as better 3D structures as well as 2D landmark predictions are learned iteratively.

\begin{figure} [ht]
  \centering
   {\vspace{-0cm}{\includegraphics[width=0.99\linewidth]{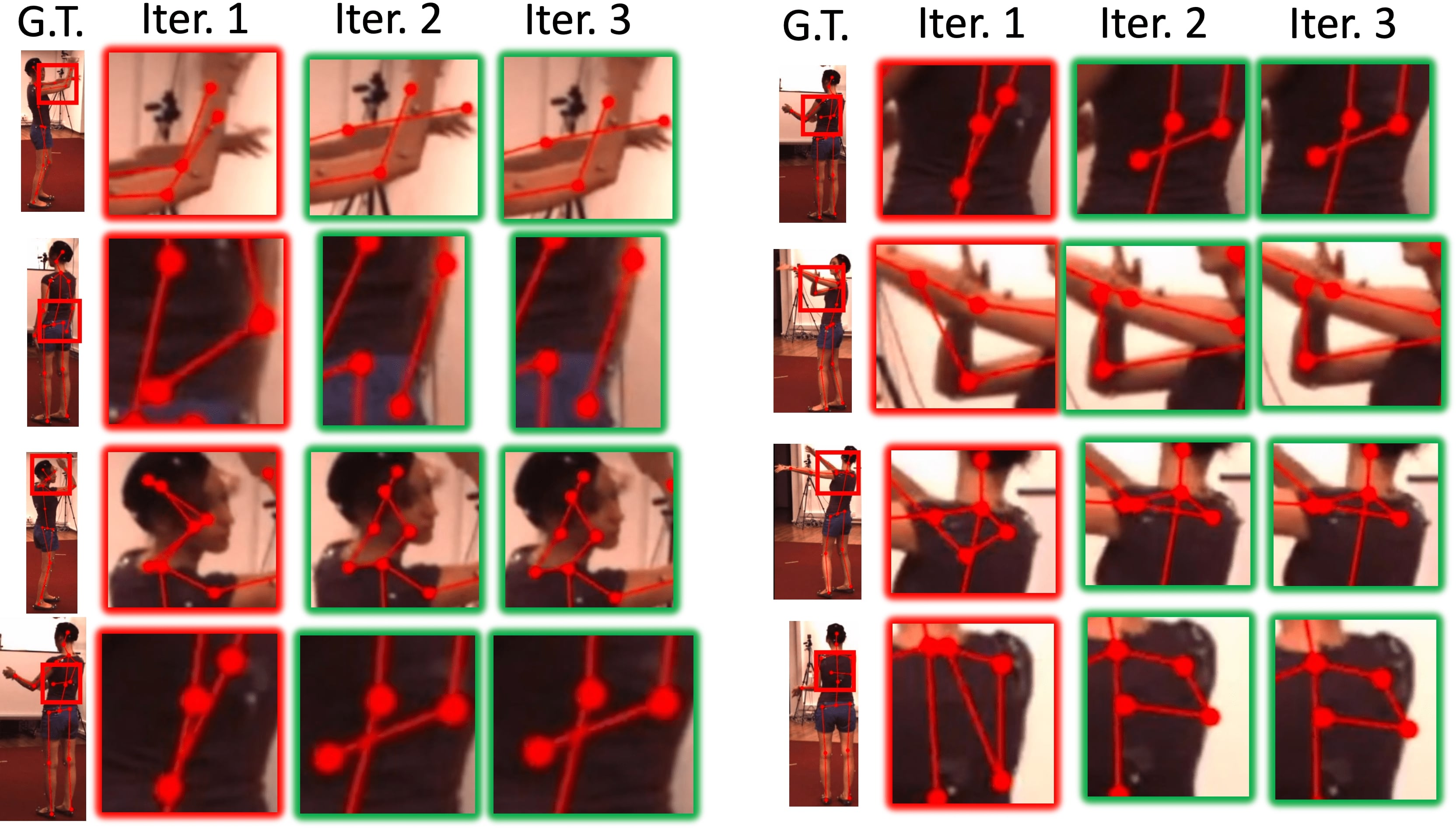} }}  
  \caption{Improvements in 2D landmark predictions as the iterations progress. Specifically, MBW is able to show improvements in cases where the landmarks are occluded. MBW leverages multi-view shape consistency from MV-NRSfM to denoise the inliers from Iter. 1 and use them as pseudo-labels for the next iteration. The red box in G.T. shows where the location of occlusion as well as groundtruth landmark locations. The red glow boxes show the noisy inliers. The green glow boxes show accurate 2D landmark predictions. This figure shows improvements during Iter. 1 }
  \label{Fig: occlusion_stage_2_improvements}
  \vspace{-0.0cm}
\end{figure}

\begin{figure} [!t]
  \centering
   \subfloat[Mean-Per-Joint-Position-Error in pixels (2D error). ]{\hspace{-0cm}{\includegraphics[width=0.55\linewidth]{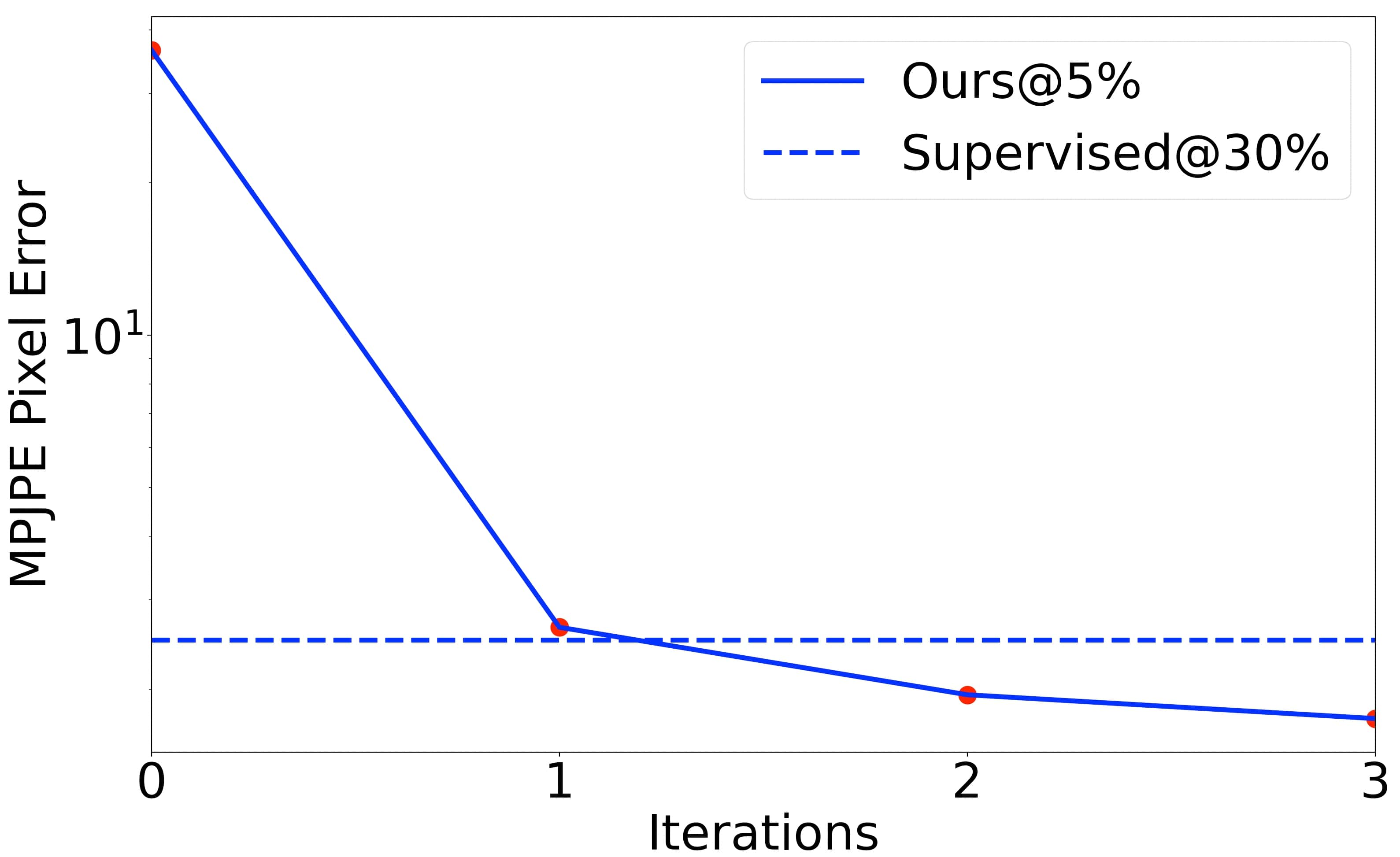} }\label{Fig: 2d_error}}  
  \subfloat[Improvements in 2D landmark prediction during Iter. 3]{\hspace{0cm}{\includegraphics[width=0.35\linewidth]{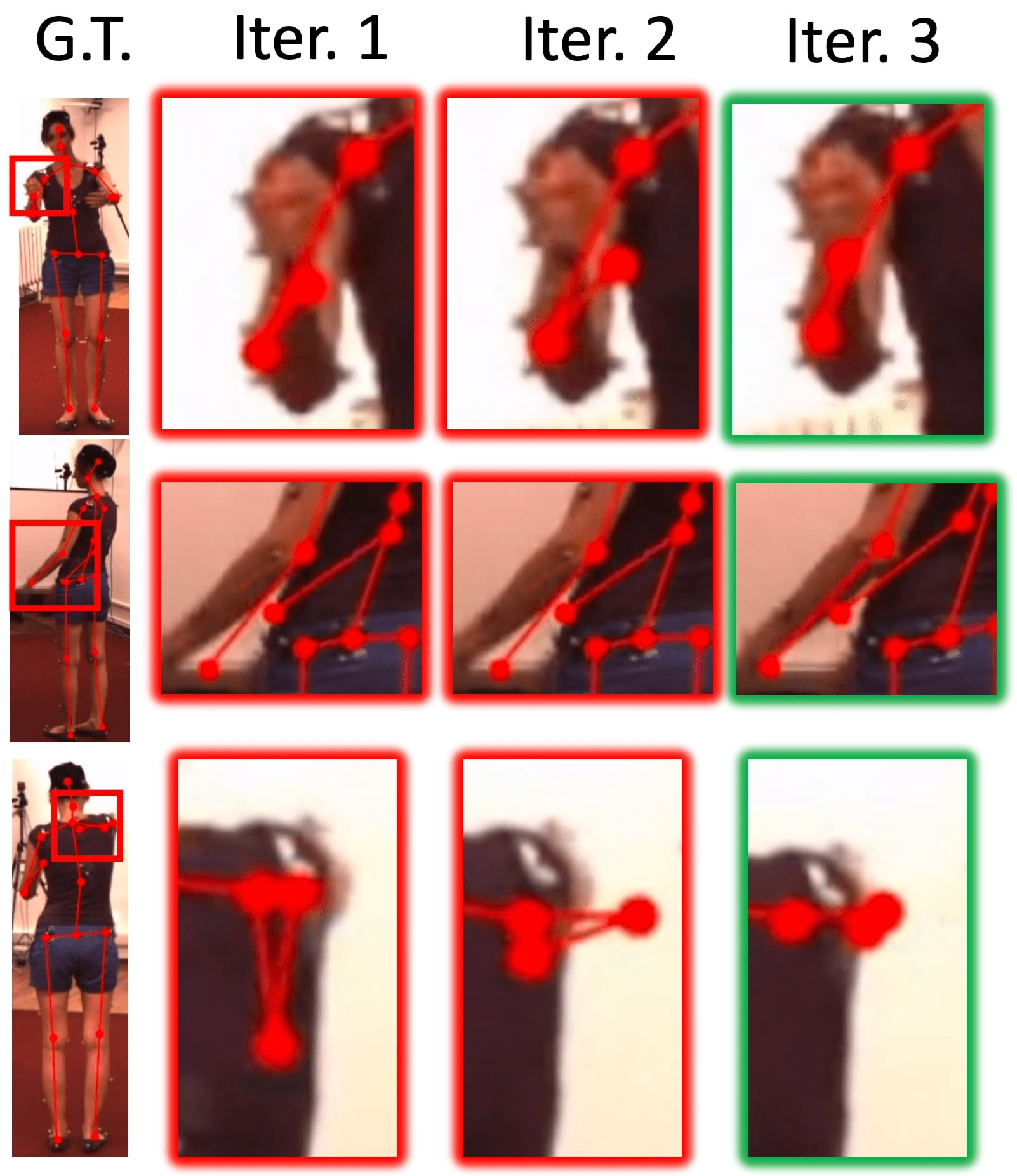} }\label{Fig: occlusion_stage_3_improvements}}
  \caption{(a) 2D landmark predictions show improvement as the iterations progress. We plot absolute errors in 2D,\textit{i.e.} we calculate Mean-Per-Joint-Position-Error in pixels for each iteration. (b) Similar to Fig.~\ref{Fig: occlusion_stage_2_improvements}, we show improvements in cases with occlusion using the proposed pipeline. This figure shows improvements during Iter. 2 }
  \label{Fig: 2d_error_iter_3}
  \vspace{-0.0cm}
\end{figure}

\textbf{Handling occlusions with geometry}:
Analyzing further, we investigate the type of improvement over different iterations. We notice that the main benefit of using learnable geometric self-supervision (see Sec. 3.2) is its capability to handle occlusions. Figure~\ref{Fig: occlusion_stage_2_improvements} shows that MBW, in conjunction with MV-NRSfM is able to denoise the 2D landmark predictions as seen in the Iter. 2 columns. Compared to Iter. 1, we observe that MV-NRSfM was able to denoise and then feed the pseudo-label to our iterative pipeline which resulted in correct annotations for cases with severe occlusions. Owing to the above observations, we show improvement of 2D landmark predictions over iterations, specifically in cases where the landmarks are occluded. Quantitative improvement is shown in Fig.~\ref{Fig: 2d_error} while the qualitative improvement is shown in Fig.~\ref{Fig: occlusion_stage_2_improvements} that shows improvements during Iter. 2, and Fig.~\ref{Fig: occlusion_stage_3_improvements} that shows improvement during Iter. 3 --  where we observe the benefit of using the multi-view constraint of MV-NRSfM.

\textbf{Denoising and its limitations}: Since MV-NRSfM leverages the redundancy in shape variations among different frames, it is less sensitive to the variations of input views, and more capable of detecting outliers and denoising inlier 2D landmark estimates. More specifically, it has the capability of denoising the 2D inputs and providing a 3D structure based on its learned distribution. For the cases shown in Fig.~\ref{Fig: occlusion_stage_2_improvements} and Fig.~\ref{Fig: occlusion_stage_3_improvements}, we showcase the denoising capabilities of MV-NRSfM. However, we should note that MV-NRSfM is only able to denoise and refine the inlier estimates if the amount of noise in 2D input labels is small enough. There are two reasons: (i) Inaccurate camera matrix: If the 2D input is extremely noisy in one of the views, even if MV-NRSfM would degenerate to an accurate 3D structure, it would not be able to reliable project the generated 3D structure over the extremely noisy view because of inaccurate camera matrix calculated from OnP or PnP. (ii) Inaccurate 3D structure: If most of the views are noisy or if the baseline between cameras is not wide enough, MV-NRSfM cannot learn to enforce multi-view shape consistency thereby generating an inaccurate 3D structure.

\begin{table} [hbt!]
\caption{Quantitative comparison showing improvement in 3D structure over each iteration on benchmark dataset~\cite{human36m}.}
  \label{Tab: 3D_over_iterations}
  \centering
  \begin{tabular}{ccc}
    \toprule
    Iteration  & 3D (mm)$\boldsymbol{\downarrow}$ \\
    \midrule
    1 & ${33.3}$ \\
    2 &  ${28.3}$ \\
    3 & $\mathbf{23.1}$ \\
    \bottomrule
  \end{tabular}
\end{table}

\section{Initial input labels and Active learning}

For the inital set of 2D input labels, we sample uniformly across time and views (unlike ~\citet{sleap} that uses PCA to decide which frames to label). Although we pick labels from each view initially, we carry this action in the initial iteration to learn a good 3D shape prior that enforces multi-view consistency. For the subsequent iterations, we do not necessarily require to pseudo labels for all the views of an instance. 

In the experiments shown in Sec. 4, we did not have a second set of manual annotations (active learning). We observe that since we propose a principled way to detect outliers, our pipeline could be readily used in the active learning domain where our proposed approach of iterations can be useful to dictate the next set of labeling in active learning. Thus, if we have a contiguous chunk of outliers in space and time for our captured video sequence, MBW could find this chunk of outliers in a principled way and is able to exactly specify where we should sample and collect more data for the next iteration if it is used for active learning.

From the above discussion, it is clear that our approach has a rich connection with active learning and could be readily used with the work from ~\citet{active_learning_meta} that uses active learning to iteratively improve the performance of the network in each iteration. Although it is outside of the scope of the paper, we would like to note that the strengths of our results show its applicability in the active learning scenario.



\section{MV-NRSfM Architecture Details}  \label{appendix: mvnrsfm_arch}

\begin{figure} [ht]
    \centering
    \includegraphics[width=0.9\linewidth]{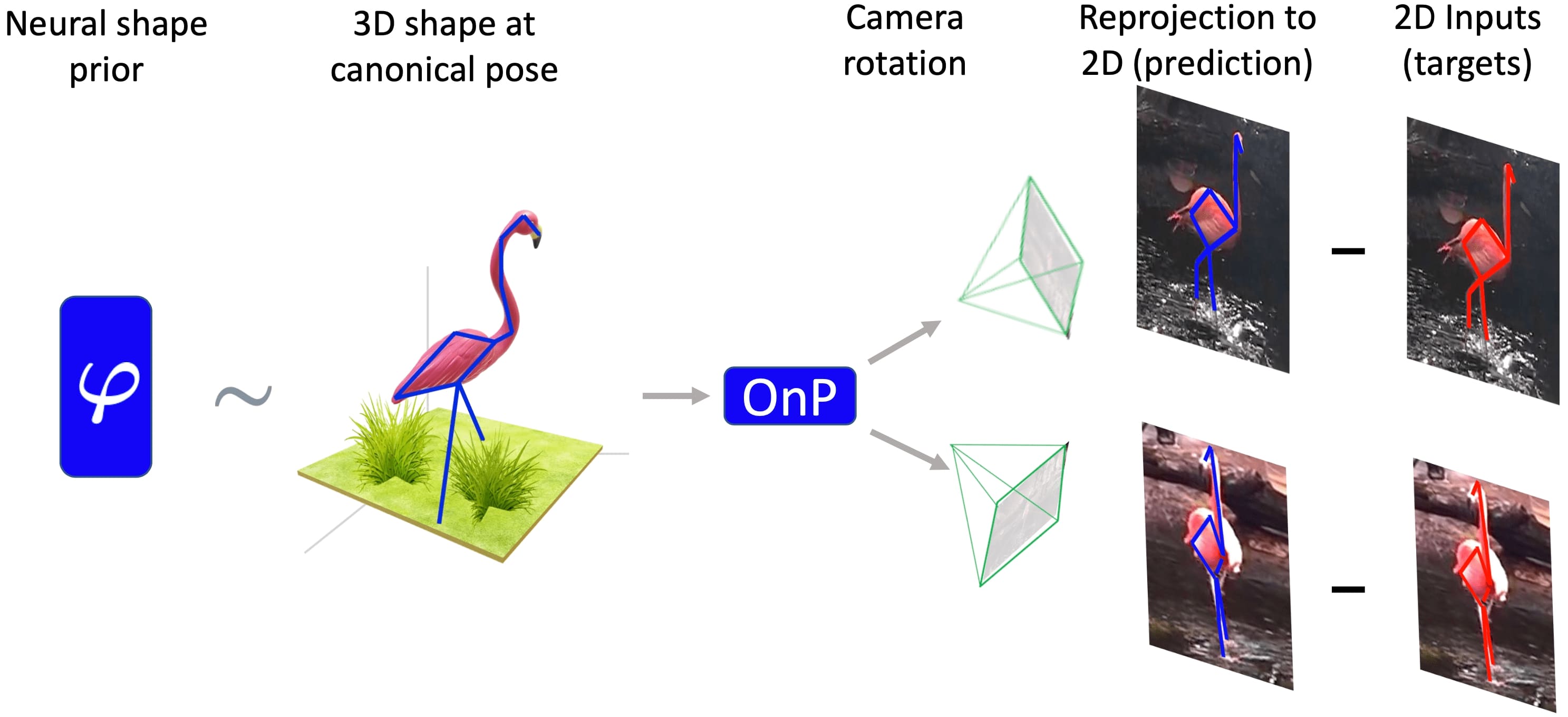}
  \caption{The 3D structure is drawn from a statistical shape distribution using neural shape priors and consequently projected to multiple views using the cameras calculated through OnP formulation. MV-NRSfM minimizes the 2D reprojection error between the predicted 2D projections and target (input) 2D projections.}
  \label{Fig: mv-nrsfm-arch}
  \vspace{-0.1cm}
\end{figure}

As shown in Fig.~\ref{Fig: mv-nrsfm-arch}, a 3D structure is drawn from a statistical shape distribution, and consequently projected to 2 or more views using a Perspective-n-Point or Orthographic-n-Point solver. Given a set of frames, the parameters of the shape distribution are adapted by minimizing the error between the predicted and the ground truth 2D keypoints. 

\begin{figure} [!t]
    \centering
    \includegraphics[width=0.9\linewidth]{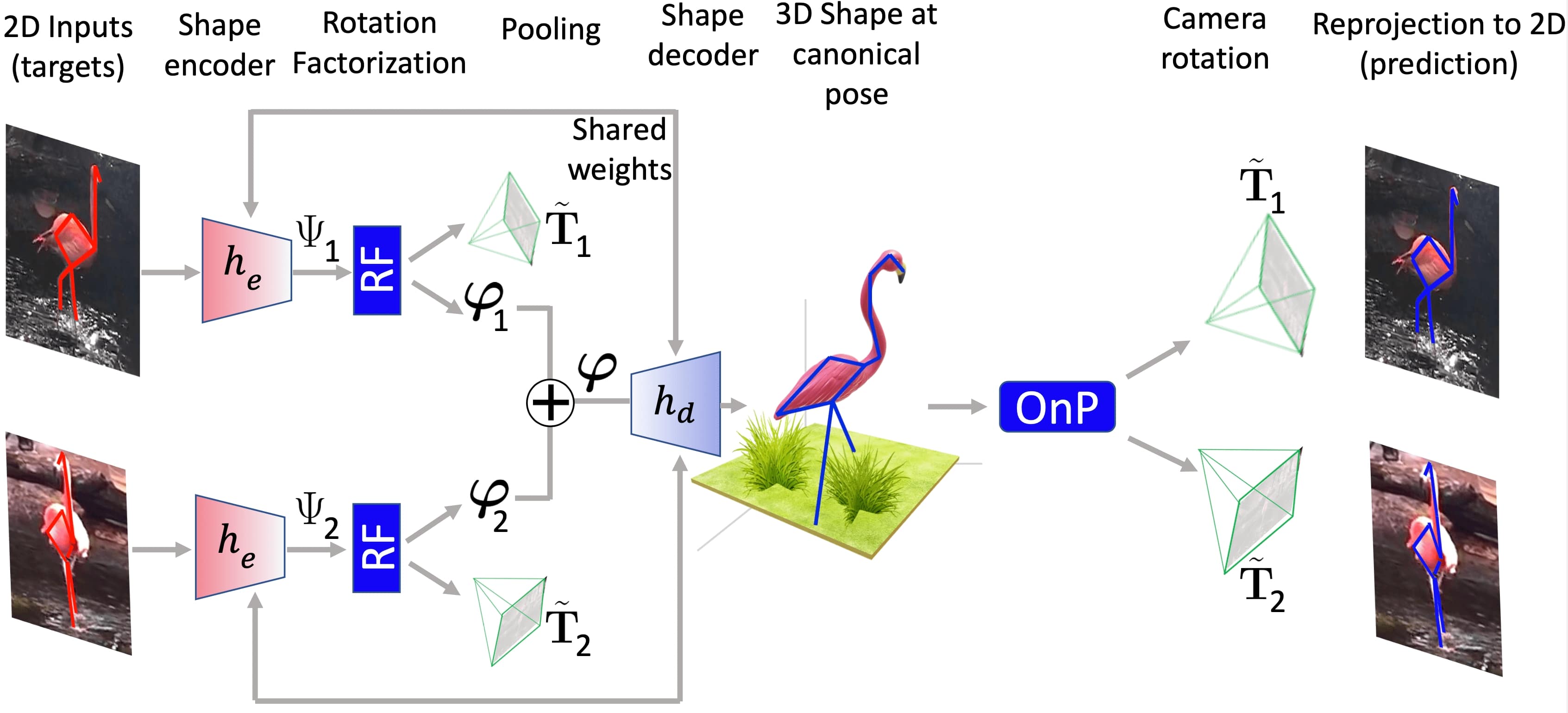}
  \caption{Architecture showing autoencoder-based MV-NRSfM approach. The 2D inputs (targets) from multiple views act as an input to encoder $\boldsymbol{h}_{e}$ that extracts the block sparse code $\Psi$ from the corresponding views. A Rotation Factorization (RF) layer at the bottleneck stage, factorizes the block sparse code into the respective camera matrices and the unrotated vector sparse code $\varphi$. The codes are then fused via \textit{pooling} function into a single code that acts as an input to the shape decoder $\boldsymbol{h}_{d}$. The shape decoder predicts the 3D structure in the canonical frame while enforcing equivariant view consistency.}
  \label{Fig: mv-nrsfm-arch-autoencoder}
  \vspace{-0.5cm}
\end{figure}

The architecture of the multi-view neural shape prior is shown in Fig.~\ref{Fig: mv-nrsfm-arch-autoencoder}. Motivated by hierarchical sparse coding~\cite{deep_nrsfm}, we implement the neural shape prior with an autoencoder with a bottleneck dimension of $8$ (we keep the same bottleneck dimension for all of our experiments across all different object categories). First, the network $h_{e}$ extracts the block sparse codes, $\Psi$. Thereafter, the bottleneck layer factorizes the block sparse code into a projection (camera) matrix and the unrotated vector sparse code, $\varphi$. We use the same encoder over all additional views. The vector sparse codes at the bottleneck stage are then pooled together and fed into a shape decoder, $h_{d}$ to generate the 3D shape at the canonical pose. Finally, the canonical 3D structure is projected over all the views, using the closed-form solution of the PnP or OnP solver.

\section{Bounding Box calculation}  \label{appendix: bbox}

\begin{figure} [!t]
  \centering
   \subfloat[Bounding box during initial iteration. ]{\vspace{-0cm}{\includegraphics[width=0.48\linewidth]{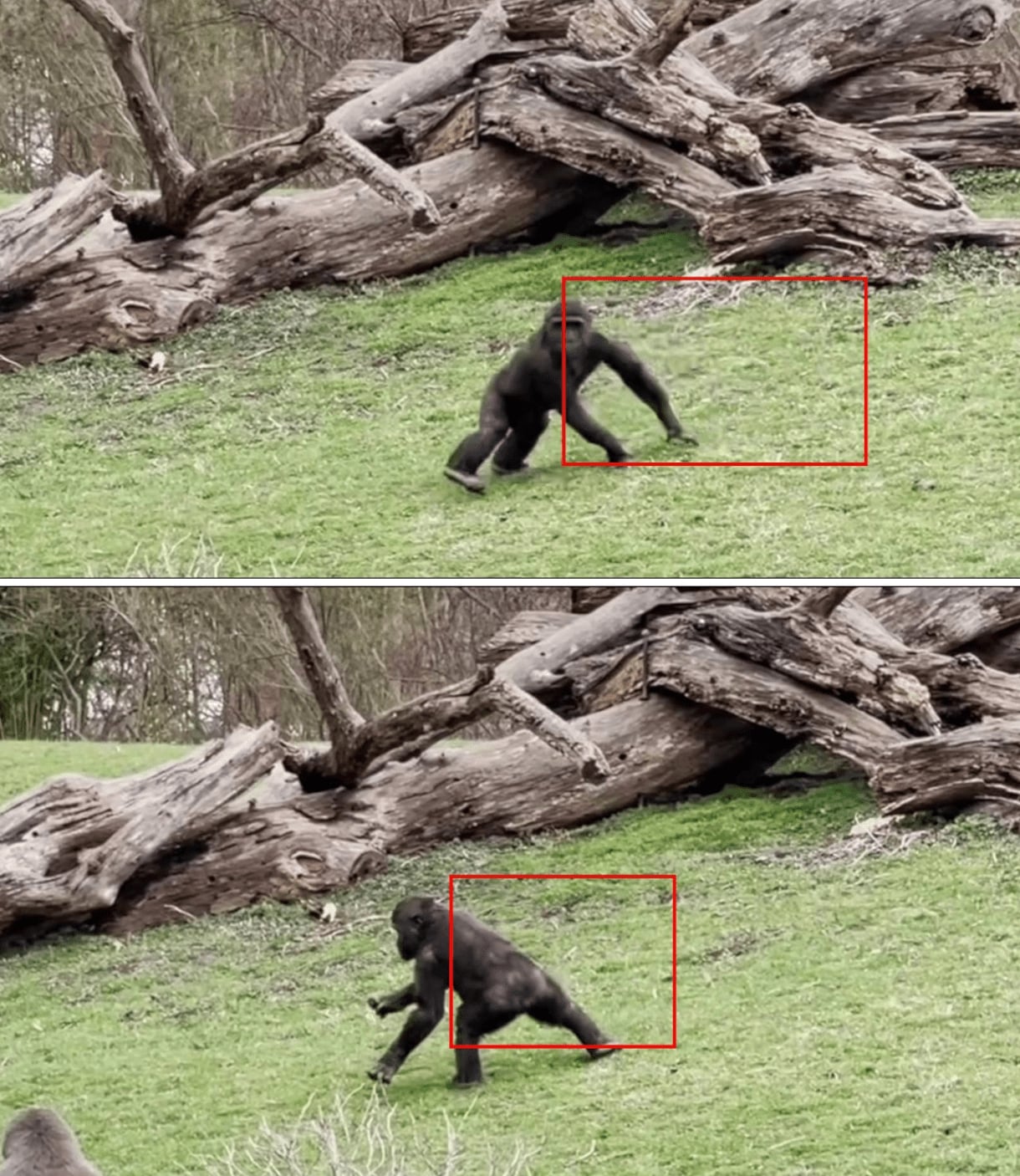} }\label{Fig: initial_bbox_instance}}  
  \subfloat[Bounding box during subsequent iterations. ]{\vspace{-0cm}{\includegraphics[width=0.48\linewidth]{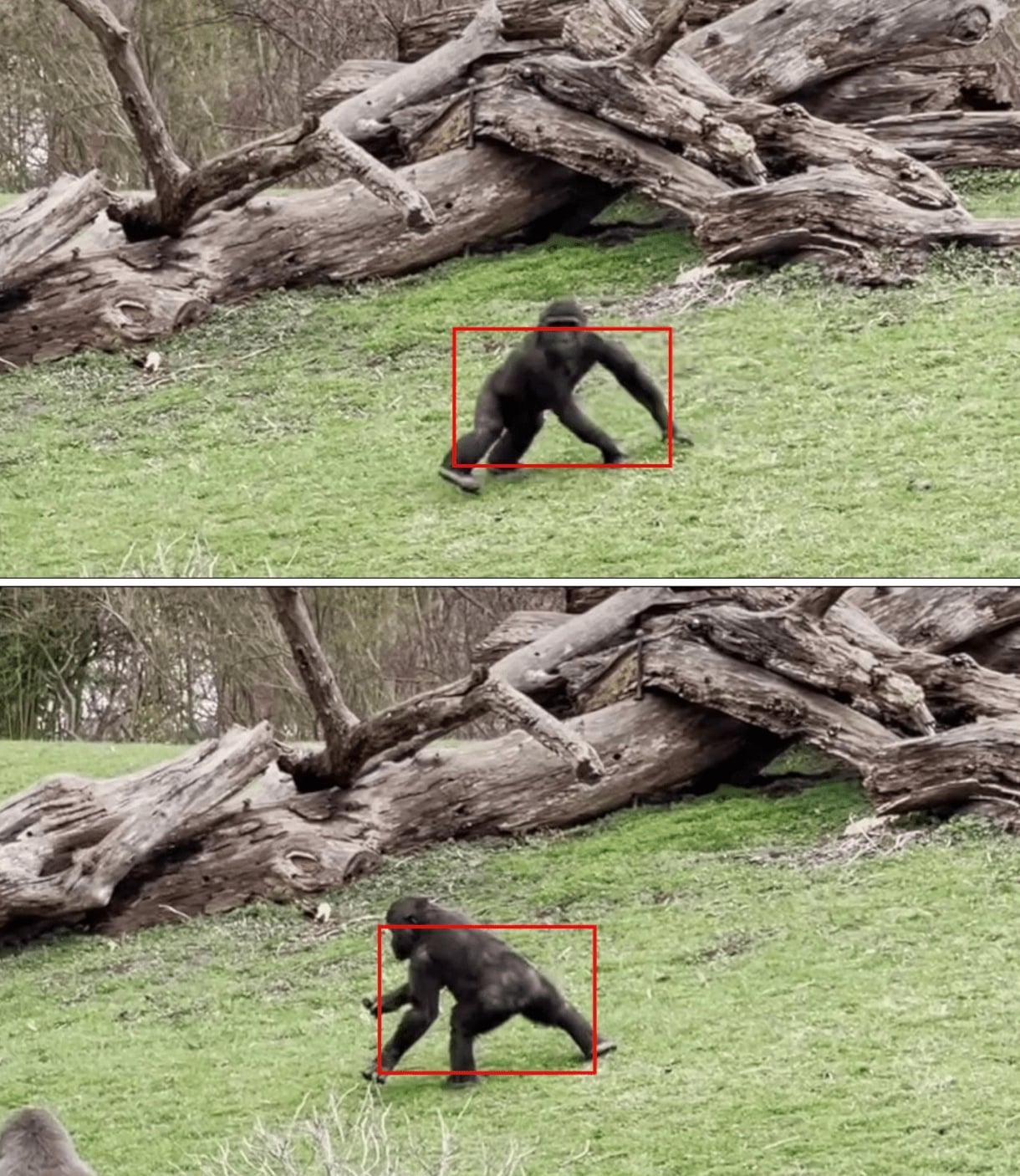} }\label{Fig: final_bbox_instance}}
  \caption{(a) Bounding box visualization during the initial iteration - calculated from the optical flow 2D landmark predictions. Unless readjusted, since optical flow predictions tend to accrue errors, we see that the calculated bounding box exhibit a similar error (elongation to the right). (b) During subsequent iterations, the bounding box predictions become accurate due to less noisy 2D candidate predictions. }
  \label{Fig: bbox_plots}
  \vspace{-0.1cm}
\end{figure}

We assume that only a single object of interest (Chimpanzee in Fig.~\ref{Fig: bbox_plots} is visible in each frame. In this section, we briefly explain our technique to estimate bounding boxes to reduce the problem into a single object. Using the initial set of 2D input labels, we propagate optical flow~\cite{raft} and generate 2D candidate predictions over the entire sequence. During this initial (flow iteration) iteration, we calculate the bounding box for the entire sequence using the 2D candidate predictions -- by taking the smallest and largest $x$ and $y$ coordinates of 2D candidate predictions. However, since the 2D candidate predictions coming from the optical flow are unreliable and noisy, we pad the bounding box with extra space by a fixed size to make the object visible, as shown in Fig.~\ref{Fig: initial_bbox_instance}. For the subsequent iterations, we calculate the bounding boxes from the 2D detector network predictions that are denoised through MV-NRSfM. Since we already denoise the 2D candidate predictions, we remove the extra padding from the bounding box calculation. An experimental video showing the bounding box visualization improvement from initial to final iteration is attached in the supplementary. 

\section{MBW-Zoo Dataset}
The URL to platform where the dataset can be viewed and downloaded:  {\fontfamily{qcr}\selectfont \textcolor{magenta}{https://github.com/mosamdabhi/MBW-Data}}. Further information concerning the released data asset such as datasheets for dataset, and overall dataset documentation is discussed in the following subsections.

\subsection{Overview} \label{sec: data_motive}

We release a challenging dataset and consisting image frames of tail-end distribution categories (such as Fish, Colobus Monkeys, Chimpanzees, etc.) with their corresponding 2D, 3D, and Bounding-Box labels generated from minimal human intervention. Some of the prominent use cases of this dataset include not only sparse 2D and 3D landmark prediction, but also dense reconstruction tasks such as dense deformable shape reconstruction, novel view rendering (NeRF), and finally this dataset could also be used for advancing Simultaneous Localization and Mapping (SLAM) frameworks. We also submit the codebase that we used to generate the above labels for in-the-wild object categories.

The data was collected by two smartphone cameras without any constraints: meaning no guidance or instructions were given as to how the data should be collected. The intention was to mimic the data captured casually by anyone holding a smartphone grade camera. Due to this reason, the cameras were continuously moving in space changing their extrinsics with respect to each other, capturing an in-the-wild dynamic scene. Thus, this dataset could be used to benchmark robust algorithms in various computer vision tasks.

\begin{figure} [!h]
    \centering
    \includegraphics[width=1\linewidth]{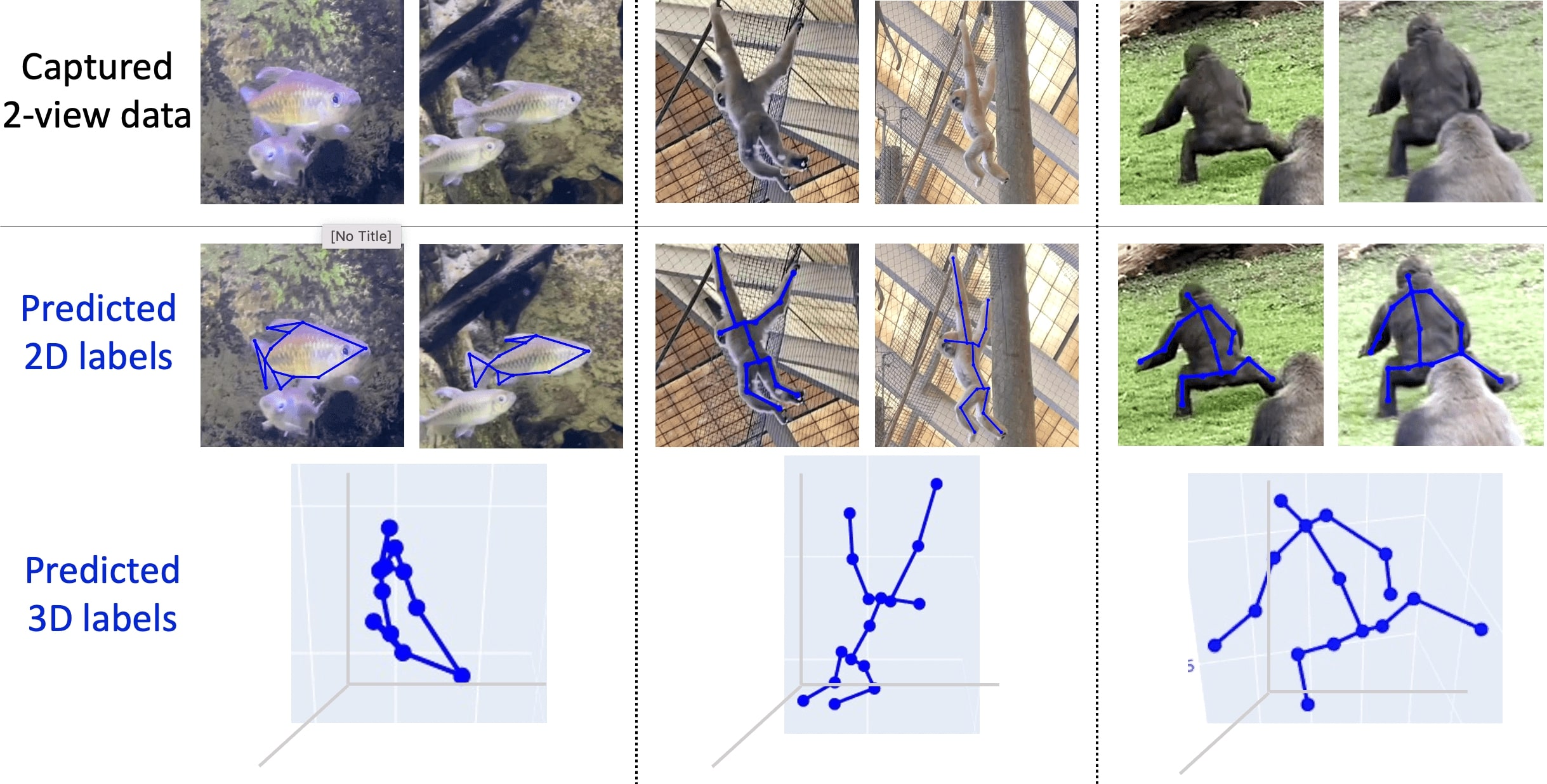}
  \caption{Overview of the dataset. The dataset consists of 2-view synchronized video sequences captured from a zoo visit. MBW provides landmark predictions (labels) that we provide for the categories shown above.}
  \label{Fig: overview}
  \vspace{-0.0cm}
\end{figure}

\subsection{Collection process}

We capture 2-View videos from handheld smartphone cameras. In our case, we used an iPhone 11 Pro Max and an iPhone 12 Pro Max to capture the video sequences. We use Final Cut Pro to manually synchronize the 2-View video sequences using the audio signal and time stamps. Please note that all we require are 2-view synchronized image frames and manual annotations for 1-2\% of the data. No camera calibration (intrinsics or extrinsics) is required to run MBW. 

\subsection{Frames visualization}

\begin{figure} [!h]
    \centering
    \includegraphics[width=1\linewidth]{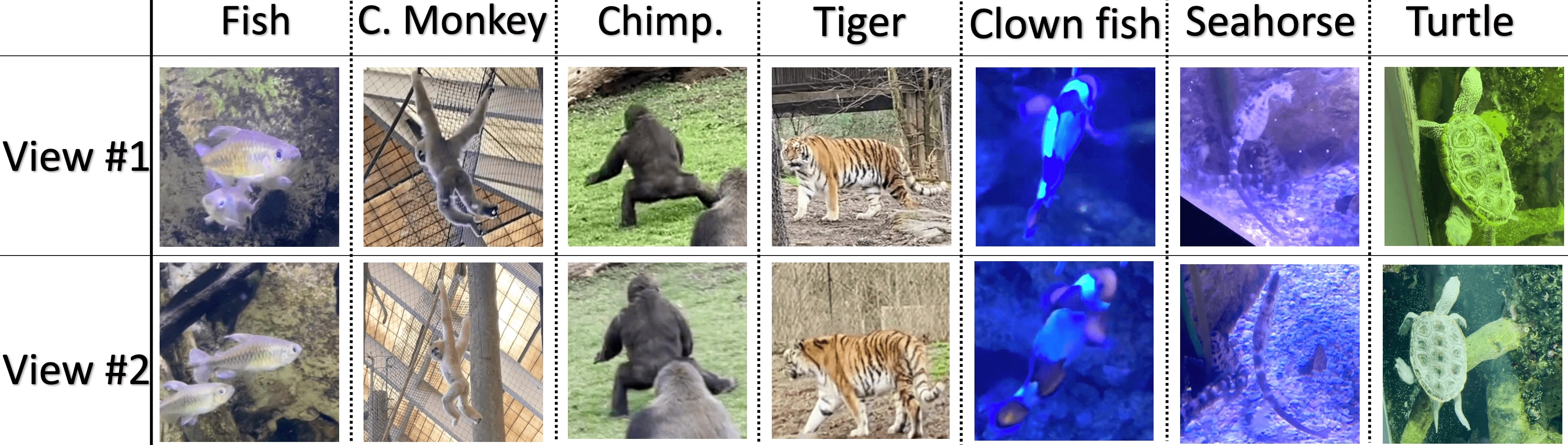}
  \caption{Visualization of instances from 7 different object categories.}
  \label{Fig: instances}
  \vspace{-0.0cm}
\end{figure}

In total, there are 16154 instances in this dataset from 7 different object categories, coming from 2 camera views. Sample instances are visualized in Fig.~\ref{Fig: instances}.

\subsection{Joint connections visualization}

We manually annotate 1-2\% of the image frames per view. Our annotation consists of the 2D landmark keypoints. The location of landmarks is chosen to extract articulated information from the objects. Joint connection visualization is shown below.

\begin{figure} [!h]
    \centering
    \includegraphics[width=0.9\linewidth]{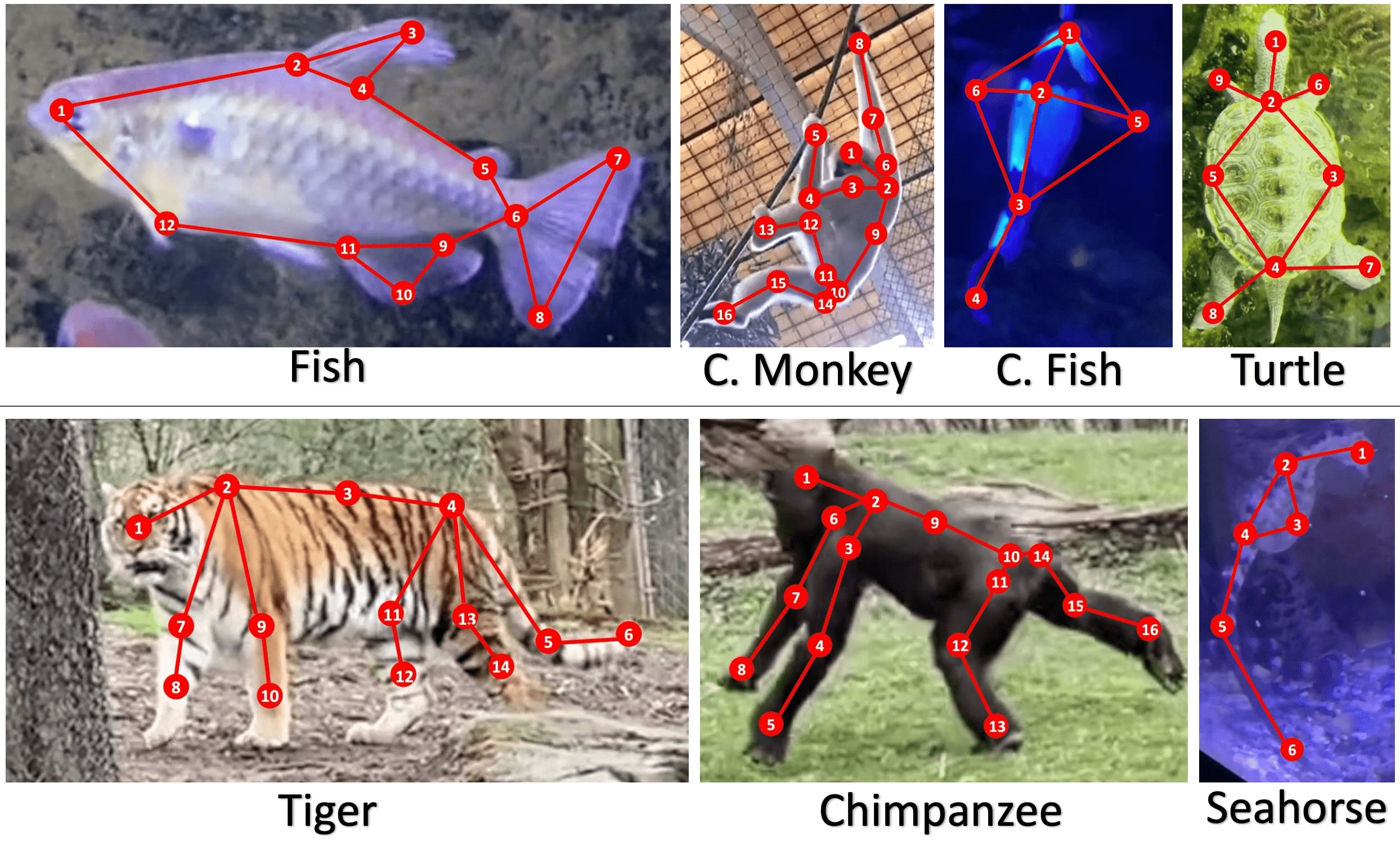}
  \caption{Visualization of keypoint connections on different object categories.}
  \label{Fig: joint_connections}
  \vspace{-0.0cm}
\end{figure}

\paragraph{How to choose the number of joints to track?}
Different applications require the tracking of different keypoints (landmarks/joints) of tail-end distribution objects. Thus, our approach (MBW) gives user the freedom to decide which keypoints are of interest and hence should be chosen to track. In the released dataset, we chose keypoints that explain the articulation or deformation of the object catgeory, that we visualize the in Fig.~\ref{Fig: joint_connections}.

\section{Dataset format} \label{sec: dataset_format}

The dataset (can be viewed and downloaded by the reviewers from here~\cite{Dabhi_MBW_Zoo_data}. The downloaded dataset is divided into two directories: {\fontfamily{qcr}\selectfont \textcolor{magenta} {annot/}} and {\fontfamily{qcr}\selectfont \textcolor{magenta} {images/}}. As names suggest, the {\fontfamily{qcr}\selectfont \textcolor{magenta} {annot/}} directory contains annotations and {\fontfamily{qcr}\selectfont \textcolor{magenta} {images/}} directory consists of 2-view synchronized image frames.

\paragraph{Annotations format}
The annotation format is discussed in Tab.~\ref{Tab: annotation_format}.

\begin{table} [!h]
\caption{The annotations are provided as a {\fontfamily{qcr}\selectfont \textcolor{magenta} {.pkl}} file. The pickle files consists of following keys.}
  \label{Tab: annotation_format}
  \centering
  \begin{tabular}{ll}
    \toprule
    Key    & Description \\
    \midrule
    {\fontfamily{qcr}\selectfont \textcolor{magenta} {W\_GT}}  & Manual annotation. Non-NaN values for $\approx 2$\% of data. NaN values for the rest.   \\    
    
    {\fontfamily{qcr}\selectfont \textcolor{magenta} {W\_Predictions}}  & 2D landmark predictions (labels) generated from MBW.   \\    
    
    {\fontfamily{qcr}\selectfont \textcolor{magenta} {S\_Pred}}  & 3D landmark predictions (labels) generated from MBW (up-to-scale).    \\    

    {\fontfamily{qcr}\selectfont \textcolor{magenta} {BBox}}  & Bounding box crops generated from MBW  \\    
    
    {\fontfamily{qcr}\selectfont \textcolor{magenta} {confidence}}  & Flag specifying confidence (Eq.~\eqref{eq:uncertainty_score}) for the MBW predictions.   \\

    \bottomrule
  \end{tabular}
\end{table}

\section{Datasheets for dataset}

As part of making dataset collection more easy and amenable in a wildly unconstrained setup, we collect a dataset of zoo animals using smartphone grade cameras, and annotate ($\approx 2$\%) of the collected frames with 2D keypoint landmarks. We call this dataset the Multiview Bootstrapping in the wild (MBW) Zoo dataset; what follows below is the datasheet~\cite{datasheets} describing this data. 

\subsection{Motivation}

\begin{enumerate}
    \item \textbf{For what purpose was the dataset created?} \textit{(Was there a specific task in mind? Was there a specific gap that needed to be filled? Please provide a description.)}

    \paragraph{}     This dataset was created to generate 2D and 3D labels of articulated objects in unconstrained settings. Such unconstrained and casually collected datasets have a wide variety of applications including entertainment, neuroscience, psychology, ethology, and many fields of medicine. Large offline labeled datasets do not exist for all but the most common articulated object categories (e.g., humans, hands, cars). Hand labeling these landmarks within a video sequence is a laborious task. Learned landmark detectors can help, but can be error-prone when trained from only a few examples. As part of contribution of this paper, we provide this dataset where 2D and 3D labels are generated in very challenging scenarios, using our approach. 
    
    Note that the user is required to only provide handheld videos from 2 or more views and manual 2D keypoint labels for ~15 frames per video. No camera intrinsics or extrinsics information is required to generate the labels shown in this dataset.
    
    \vspace{0.25cm}
    \item \textbf{Who created this dataset?} \textit{ (e.g., which team, research group) and on behalf of which entity (e.g., company, institution, organization)}?
    
    \paragraph{} This dataset was created by the corresponding author, Mosam Dabhi. At the time of creation, Mosam is a graduate student at Carnegie Mellon University (CMU) in Pittsburgh, Pennsylvania, USA.
    
    \vspace{0.25cm}
    \item \textbf{Who funded the creation of the dataset?} \textit{(If there is an associated grant, please provide the name of the grant or and the grant name and number.)}
    
    \paragraph{} 
    \answerNA
    
    \vspace{0.25cm}
    \item \textbf{Any other comments?}
    
    \paragraph{}\answerNA
\end{enumerate}

\subsection{Composition} \label{sec: datasheet_composition}
\begin{enumerate}
    \item \textbf{What do the instances that comprise the dataset represent} \textit{(e.g., documents, photos, people, countries)}?
    
    \paragraph{} Each instance is an image of an articulated object (zoo animals, birds, and fish). Approximately 2-5\% of images have corresponding manual 2D landmark annotation. From here on, we use the term landmark prediction and label interchangeably for convenience. For the remaining images in the sequences, the 2D and 3D landmark labels, as well as bounding box crops labels are generated by MBW. In particular, each entity also has a confidence flag that is a boolean specifying how much reliable is the label generated by MBW.
    
    \vspace{0.25cm}
    \item \textbf{How many instances are there in total (of each type, if appropriate)?} \label{ques: composition_2}

    \paragraph{} In total, there are 16154 instances in this dataset from 7 different object categories, coming from 2 camera views. The overall dataset statistics in Tab.~\ref{Tab: dataset_specs} reflect the above description.

\begin{table} [!h]
\caption{Dataset composition specifications.}
  \label{Tab: dataset_specs}
  \centering
  \begin{tabular}{ccccc}
    \toprule
    Object    & Frames (\#)    & Joints (\#) &  Manual labels (\%) & Labels from MBW \\
    \midrule
    Fish & $1456$ &  12 & $2.7$ & \textbf{Available} \\
    Colobus Monkey & $392$ &  $16$ & $5.1$ &  \textbf{Available} \\
    Chimpanzee & $204$ &  $16$ & $6.3$ &  \textbf{Available} \\    
    Tiger & $1829$ &  $14$ & $0.4$ & \textbf{Available} \\        
    Clownfish & $910$ &  $6$ & $2.0$ & \textbf{Available} \\        
    Seahorse & $480$ &  $6$ & $2.2$ & \textbf{Available} \\        
    Turtle & $2806$ &  $9$ & \answerNA & \answerNA \\        
    \bottomrule
  \end{tabular}
\end{table}   

    \paragraph{Note:}  We are unable to provide the stereo baseline distance (m) and stereo angle ($^{\circ}$) since the data was captured where the cameras were continuously moving thereby changing these metrics. 

    \vspace{0.25cm}
    \item \textbf{Does the dataset contain all possible instances or is it a sample (not necessarily random) of instances from a larger set?} \textit{ (If the dataset is a sample, then what is the larger set? Is the sample representative of the larger set (e.g., geographic coverage)? If so, please describe how this representativeness was validated/verified. If it is not representative of the larger set, please describe why not (e.g., to cover a more diverse range of instances, because instances were withheld or unavailable)?} \label{ques: composition_3}
    
    \paragraph{} It is a sample of all videos captured casually in an unconstrained environment such as zoo. It is not intended to be representative: the data was collected randomly in the order of visit. This data was collected with the intent to show the applicability of MBW in challenging data scenarios -- specifically to label articulated objects in the wild at scale.

    \vspace{0.25cm}
    \item \textbf{What data does each instance consist of?} \textit{(``Raw'' data (e.g., unprocessed text or images)or features? In either case, please provide a description)}.
    
    \paragraph{} Please refer Sec.~\ref{sec: dataset_format} and the corresponding Table~\ref{Tab: annotation_format}.
    
    \vspace{0.25cm}
    \item \textbf{Is there a label or target associated with each instance? If so, please provide a description.}
    
    \paragraph{} As noted in the Table~\ref{Tab: annotation_format}, labels for each instance are associated for the categories with flag \textbf{Available}. For the rest, only initial $\approx 2$\% {\fontfamily{qcr}\selectfont \textcolor{magenta} {W\_GT}} labels are provided, since we did not run MBW that could provide us with prediction labels. We release this dataset to set a benchmark for solving challenging 2D and 3D landmark prediction tasks for in-the-wild unconstrained video captures.
    
    \vspace{0.25cm}
    \item \textbf{Is any information missing from individual instances?} \textit{(If so, please provide a description, explaining why this information is missing (e.g., because it was unavailable). This does not include intentionally removed information, but might include, e.g., redacted text.)*}
    
    \paragraph{} The prediction labels ({\fontfamily{qcr}\selectfont \textcolor{magenta} {W\_Predictions}} , {\fontfamily{qcr}\selectfont \textcolor{magenta} {S\_Pred}} ,
    {\fontfamily{qcr}\selectfont \textcolor{magenta} {BBox}}, and {\fontfamily{qcr}\selectfont \textcolor{magenta} {confidence}} ) are missing for the categories where MBW is not run as shown in Table~\ref{Tab: annotation_format}.
    
    \vspace{0.25cm}
    \item \textbf{Are relationships between individual instances made explicit? (e.g., users' movie ratings, social network links)?} \textit{If so, please describe how these relationships are made explicit.}
    
    \paragraph{} Instances are unrelated.
    
    \vspace{0.25cm}
    \item \textbf{Are there recommended data splits (e.g., training, development/validation, testing)?} \textit{If so, please provide a description of these splits, explaining the rationale behind them.}
    
    \paragraph{} Since the sole purpose of this data collection was to generate labels from scratch, we expect this data to be used solely for generating labels for unlabeled data. Thus, we do not explicitly provide a training/validation/testing split; however, we recognize that people may wish to do this, or to do some form of cross-validation. We would suggest cross-validation and test split by dividing the manual labels into 80/10/10 split and pick the samples via a uniform sampling strategy.
    
    \vspace{0.25cm}
    
    \item \textbf{Are there any errors, sources of noise, or redundancies in the dataset?} \textit{If so, please provide a description.}
    
    \paragraph{} Since the inital 2D keypoints were manually labeled, there could be some errors in the manual annotations since they were visually localized and clicked as labels.
    
    \vspace{0.25cm}
    \item \textbf{Is the dataset self-contained, or does it link to or otherwise rely on external resources (e.g., websites, tweets, other datasets)?} \textit{If it links to or relies on external resources, a) are there guarantees that they will exist, and remain constant, over time; b) are there official archival versions of the complete dataset (i.e., including the external resources as they existed at the time the dataset was created); c) are there any restrictions (e.g., licenses, fees) associated with any of the external resources that might apply to a future user? Please provide descriptions of all external resources and any restrictions associated with them, as well as links or other access points, as appropriate.)}
    
    The dataset needs to be downloaded from here~\cite{Dabhi_MBW_Zoo_data}.
    
    \vspace{0.25cm}
    \item \textbf{Does the dataset contain data that might be considered confidential (e.g., data that is protected by legal privilege or by doctor-patient confidentiality, data that includes the content of individuals' non-public communications)} \textit{If so, please provide a description.}
    
    \paragraph{} No.
    
    \vspace{0.25cm}
    \item \textbf{Does the dataset contain data that, if viewed directly, might be offensive, insulting, threatening, or might otherwise cause anxiety?} \textit{If so, please describe why.}
    
    \paragraph{} No.
    
    \vspace{0.25cm}
    \item \textbf{Does the dataset relate to people?} \textit{If not, you may skip the remaining questions in this section.}
    
    \paragraph{} No

    \vspace{0.25cm}
    \item \textbf{Does the dataset identify any subpopulations (e.g., by age, gender)?} \textit{(If so, please describe how these subpopulations are identified and provide a description of their respective distributions within the dataset.)} 
    
    \paragraph{} \answerNA

    \vspace{0.25cm}
    \item \textbf{Is it possible to identify individuals (i.e., one or more natural persons), either directly or indirectly (i.e., in combination with other data) from the dataset?} \textit{(If so, please describe how)} 
    
    \paragraph{} \answerNA    

    \vspace{0.25cm}
    \item \textbf{Does the dataset contain data that might be considered sensitive in any way (e.g., data that reveals racial or ethnic origins, sexual orientations, religious beliefs, political opinions or union memberships, or locations; financial or health data; biometric or genetic data; forms of government identification, such as social security numbers; criminal history)?} \textit{(If so, please provide a description.)} 
    
    \paragraph{} \answerNA    

    \vspace{0.25cm}
    \item \textbf{Any other comments?} 
    
    \paragraph{} \answerNA  
    
\end{enumerate}

\subsection{Collection Process} \label{sec: collection_process}
\begin{enumerate}
    \item \textbf{How was the data associated with each instance acquired?} \textit{(Was the data directly observable (e.g., raw text, movie ratings), reported by subjects (e.g., survey responses), or indirectly inferred/derived from other data (e.g., part-of-speech tags, model-based guesses for age or language)? If data was reported by subjects or indirectly inferred/derived from other data, was the data validated/verified? If so, please describe how.)} \label{ques: ques_1_collection}

    \paragraph{} The data was collected by two smartphone cameras without any constraints: meaning no guidance or instructions were given as to how the data was collected. The intention was to mimic the data captured casually by anyone holding a smartphone grade camera. Due to this reason, the cameras were continuously moving in space changing their extrinsics with respect to each other thereby making this a dynamic setup.
    
    \vspace{0.25cm}
    \item \textbf{What mechanisms or procedures were used to collect the data (e.g., hardware apparatus or sensor, manual human curation, software program, software API)} \textit{How were these mechanisms or procedures validated?}
    
    \paragraph{} We captured 2-View videos from handheld smartphone cameras. In our case, we used an iPhone 11 Pro Max and an iPhone 12 Pro Max to capture the video sequences at ~30 frames-per-second (fps). We use Final Cut Pro to manually synchronize the 2-View video sequences using the audio signal and time stamps. Please note that all we require are 2-view synchronized image frames and manual annotations for 1-2\% of the data. As mentioned above, we are unable to provide the stereo baseline distance and stereo angles since the data was captured where the cameras were continuously moving thereby changing these metrics.
    
    \vspace{0.25cm}
    \item \textbf{If the dataset is a sample from a larger set, what was the sampling strategy (e.g., deterministic, probabilistic with specific sampling probabilities)?}
    
    \paragraph{} Please refer question \#~\ref{ques: composition_2} and question\#~\ref{ques: composition_3} of Sec.~\ref{sec: datasheet_composition}.
    
    \vspace{0.25cm}
    \item \textbf{Who was involved in the data collection process (e.g., students, crowdworkers, contractors) and how were they compensated (e.g., how much were crowdworkers paid)?}
    
    \paragraph{} The authors were helped by Shraddha Thakkar who graciously volunteered to capture the data during their visit to a Zoo with authors.
    
    \vspace{0.25cm}
    \item \textbf{Over what timeframe was the data collected?} (\textit{Does this timeframe match the creation timeframe of the data associated with the instances (e.g., recent crawl of old news articles?  If not, please describe the timeframe in which the data associated with the instances was created.)}

    The dataset was collected on March 19, 2022.

    \vspace{0.25cm}
    \item \textbf{Were any ethical review processes conducted (e.g., by an institutional review board)?} \textit{If so, please provide a description of these review processes, including the outcomes, as well as a link or other access point to any supporting documentation.}
    
    \paragraph{} No review processes were conducted with respect to the collection of this data. Manual annotation was conducted by visually localizing the joints on the objects whose accuracy was confirmed by visual inspection. 
    
    \item \textbf{Does the dataset relate to people?} (\textit{If not, you may skip the remaining questions in this section.})
    
    \paragraph{} No.
    
    \vspace{0.25cm}
    \item \textbf{Did you collect the data from the individuals in question directly, or obtain it via third parties or other sources (e.g., websites)?}
    
    \paragraph{} \answerNA.
    
    \vspace{0.25cm}
    \item \textbf{Were the individuals in question notified about the data collection?} (\textit{If so, please describe or show with screenshots or other information how notice was provided, and provide a link or other access point to, or otherwise reproduce, the exact language of the notification itself.})
    
    \paragraph{} \answerNA.
    
    \vspace{0.25cm}
    \item \textbf{Did the individuals in question consent to the collection and use of their data?} \textit{(If so, please describe (or show with screenshots or other information) how consent was requested and provided, and provide a link or other access point to, or otherwise reproduce, the exact language to which the individuals consented.}
    
    \paragraph{} \answerNA.
    
    \vspace{0.25cm}
    \item \textbf{If consent was obtained, were the consenting individuals provided with a mechanism to revoke their consent in the future or for certain uses?} \textit{(If so, please provide a description, as well as a link or other access point to the mechanism (if appropriate).}
    
    \paragraph{} \answerNA.
    
    \vspace{0.25cm}
    \item \textbf{Has an analysis of the potential impact of the dataset and its use on data subjects (e.g., a data protection impact analysis) been conducted?} (\textit{If so, please provide a description of this analysis, including the outcomes, as well as a link or other access point to any supporting documentation.})
    
    \paragraph{} \answerNA.
    
    \vspace{0.25cm}
    \item \textbf{Any other comments.}
    
    \paragraph{} No.

\end{enumerate}

\subsection{Preprocessing/cleaning/labeling}
\begin{enumerate}
    \item \textbf{Was any preprocessing/cleaning/labeling of the data done (e.g., discretization or bucketing, tokenization, part-of-speech tagging, SIFT feature extraction, removal of instances, processing of missing values)?} \textit{(If so, please provide a description. If not, you may skip the remainder of the questions in this section.)} \label{ques: preproces_1}
    
    \paragraph{} We did not do any specific preprocessing or cleaning of the data except what is mentioned in question ~\ref{ques: ques_1_collection} of Sec.~\ref{sec: collection_process}. Manual annotations were labeled by visually localizing the joints on the objects in the limited (1-2\%) image frames.
    
    \vspace{0.25cm}
    \item \textbf{Was the ``raw'' data saved in addition to the preprocessed/cleaned/labeled data (e.g., to support unanticipated future uses)?} \textit{(If so, please provide a link or other access point to the ``raw'' data.)}
    
    \paragraph{} Yes, the image frames released under the {\fontfamily{qcr}\selectfont \textcolor{magenta} {images/}} directory are sampled from the original ``raw'' video.
    
    \vspace{0.25cm}
    \item \textbf{Is the software used to preprocess/clean/label the instances available?} (\textit{If so, please provide a link or other access point.})
    
    \paragraph{} Yes. We used open-source Matplotlib library ({\fontfamily{qcr}\selectfont \textcolor{magenta} {https://matplotlib.org}}) to visualize and label the joints. For further details, please refer question \#~\ref{ques: preproces_1} of this subsection.
    
    \vspace{0.25cm}
    \item \textbf{Any other comments?}
    
    \paragraph{} None.
    
\end{enumerate}

\subsection{Uses}
\begin{enumerate}
    \item \textbf{Has the dataset been used for any tasks already?} \textit{(If so, please provide a description.)}
    
    \paragraph{} Yes, the dataset has already been used for the task of 2D and 3D landmark predictions (sparse keypoints) by MBW. The task specifications are discussed in the MBW paper.
    
    \vspace{0.25cm}
    \item \textbf{Is there a repository that links to any or all papers or systems that use the dataset?} \textit{(If so, please provide a link or other access point.)}
    
    \paragraph{} No.
    
    \vspace{0.25cm}
    \item \textbf{What (other) tasks could the dataset be used for?} \label{ques: uses}
    
    \paragraph{} This dataset could be used for the following computer vision tasks:
    \begin{itemize}
        \item Dense 3D reconstruction of the given articulated object categories~\cite{dense_nrsfm}.
        \item Scene flow and optical flow generation tasks.
        \item Novel view rendering - owing to synchronized multi-view video sequences (NeRF)~\cite{nerf}.
        \item Estimation of cameras in space to aid the applications of robotics, like approaches in Simulataneous Localization and Mapping (SLAM).
    \end{itemize}

    \vspace{0.25cm}
    \item \textbf{Is there anything about the composition of the dataset or the way it was collected and preprocessed/cleaned/labeled that might impact future uses?} \textit{(For example, is there anything that a future user might need to know to avoid uses that could result in unfair treatment of individuals or groups (e.g., stereotyping, quality of service issues) or other undesirable harms (e.g., financial harms, legal risks)  If so, please provide a description. Is there anything a future user could do to mitigate these undesirable harms?)}
    
    \paragraph{} No, to the best of our knowledge.
    
    \vspace{0.25cm}
    \item \textbf{Are there tasks for which the dataset should not be used?} \textit{(If so, please provide a description.)}
    
    \paragraph{} Please refer question~\ref{ques: uses} of this subsection. Apart from that, our answer to this question is: No, to the best of our knowledge.
    
    \vspace{0.25cm}
    \item \textbf{Any other comments?}
    
    \paragraph{} None.
    
\end{enumerate}

\subsection{Distribution}

\begin{enumerate}
    \item \textbf{Will the dataset be distributed to third parties outside of the entity (e.g., company, institution, organization) on behalf of which the dataset was created?} \textit{(If so, please provide a description.)}
    
    \paragraph{} Yes, the dataset is freely available under \textbf{CC-BY-NC} license. 
    
    \item \textbf{How will the dataset will be distributed (e.g., tarball  on website, API, GitHub)?} \textit{(Does the dataset have a digital object identifier (DOI)?}
    
    \paragraph{} The dataset can be accessed from~\cite{Dabhi_MBW_Zoo_data}. The DOI for the dataset can be found on this GitHub page. 
    
    \vspace{0.25cm}
    \item \textbf{When will the dataset be distributed?} 
    
    \paragraph{} Please refer to the question above.
    
    \vspace{0.25cm}
    \item \textbf{Will the dataset be distributed under a copyright or other intellectual property (IP) license, and/or under applicable terms of use (ToU)?} \textit{(If so, please describe this license and/or ToU, and provide a link or other access point to, or otherwise reproduce, any relevant licensing terms or ToU, as well as any fees associated with these restrictions.)}
    
    \paragraph{} The dataset is distributed under a \textbf{CC-BY-NC} license.
    
    \vspace{0.25cm}
    \item \textbf{Have any third parties imposed IP-based or other restrictions on the data associated with the instances?} \textit{(If so, please describe these restrictions, and provide a link or other access point to, or otherwise reproduce, any relevant licensing terms, as well as any fees associated with these restrictions.)}
    
    \paragraph{} No.
    
    \vspace{0.25cm}
    \item \textbf{Do any export controls or other regulatory restrictions apply to the dataset or to individual instances?} \textit{(If so, please describe these restrictions, and provide a link or other access point to, or otherwise reproduce, any supporting documentation.)}
    
    \paragraph{} Not to our knowledge.
    
    \vspace{0.25cm}
    \item \textbf{Any other comments?}
    
    \paragraph{} No.
    
\end{enumerate}

\subsection{Maintenance}

\begin{enumerate}
    \item \textbf{Who is supporting/hosting/maintaining the dataset?}
    
    \paragraph{} The authors are maintaining and hosting the dataset information page on GitHub, while the dataset itself is hosted on Zenodo platform to have a persistent DOI.
    
    \vspace{0.25cm}
    \item \textbf{How can the owner/curator/manager of the dataset be contacted (e.g., email address)?}
    
    \paragraph{} E-mail address of the corresponding author is provided at the dataset access page~\cite{Dabhi_MBW_Zoo_data}.
    
    \vspace{0.25cm}
    
    \item \textbf{Is there an erratum?} \textit{(If so, please provide a link or other access point.)}
    
    \paragraph{} Currently, no. As errors are encountered, future versions of the dataset may be released (but will be versioned). The information to access the latest version (with updated DOI) will all be provided in the same GitHub location.
    
    \vspace{0.25cm}
    \item \textbf{ Will the dataset be updated (e.g., to correct labeling errors, add new instances, delete instances')?} \textit{(If so, please describe how often, by whom, and how updates will be communicated to users (e.g., mailing list, GitHub)?)}
    
    \paragraph{} Same as previous.
    
    \vspace{0.25cm}
    \item \textbf{If the dataset relates to people, are there applicable limits on the retention of the data associated with the instances (e.g., were individuals in question told that their data would be retained for a fixed period of time and then deleted)?} \textit{(If so, please describe these limits and explain how they will be enforced.)}
    
    \paragraph{} No.
    
    \vspace{0.25cm}
    \item \textbf{Will older versions of the dataset continue to be supported/hosted/maintained?} \textit{(If so, please describe how. If not, please describe how its obsolescence will be communicated to users.)}
    
    \paragraph{} Yes; all data will be versioned.
    
    \vspace{0.25cm}
    \item \textbf{If others want to extend/augment/build on/contribute to the dataset, is there a mechanism for them to do so?} \textit{(If so, please provide a description. Will these contributions be validated/verified? If so, please describe how. If not, why not? Is there a process for communicating/distributing these contributions to other users? If so, please provide a description.)}
    
    \paragraph{} Errors may be submitted by opening issues on GitHub. More extensive augmentations may be accepted at the authors' discretion.
    
    \vspace{0.25cm}
    \item \textbf{Any other comments?} None
    

\end{enumerate}

\end{document}